\documentclass[runningheads]{llncs}
\usepackage[title]{appendix}
\usepackage{epsfig}
\usepackage{graphicx}
\usepackage{amsmath}
\usepackage{amssymb}
\usepackage{subfig}

\usepackage[pagebackref=true,breaklinks=true,letterpaper=true,colorlinks,bookmarks=false]{hyperref}

\mainmatter
\def\ACCV20SubNumber{305}  

\title{MagGAN: High-Resolution Face Attribute Editing with Mask-Guided Generative Adversarial Network} 

\titlerunning{MagGAN} 
\authorrunning{Y. Wei et al.} 
\author{Yi Wei\inst{1} \and Zhe Gan \inst{2} \and Wenbo Li\inst{3} \and
Siwei Lyu\inst{4} \and Ming-Ching Chang\inst{1} \and Lei Zhang \inst{2} \and 
Jianfeng Gao\inst{2} \and Pengchuan Zhang\inst{2}}
\institute{University at Albany, State University of New York, USA \and
Microsoft Corporation, Redmond, USA \and Samsung Research America AI Center, USA \and
University at Buffalo, State University of New York, USA}

\usepackage{url}            
\usepackage{booktabs}       
\usepackage{amsfonts}       
\usepackage{nicefrac}       
\usepackage{microtype}      
\usepackage{caption}

\usepackage{geometry,mathtools}
\usepackage{nccmath}
\usepackage{multirow}
\usepackage{multicol}

\usepackage{amsmath}
\usepackage{amssymb, color}
\usepackage{float}

\renewcommand{\phi}{\varphi}

\renewcommand{\hat}{\widehat}

\newcommand{\etc}{\textit{etc} }
\newcommand{\ie}{\textit{i}.\textit{e}.}
\newcommand{\eg}{\textit{e}.\textit{g}.}
\newcommand{\vs}{\textit{v}.\textit{s}.}

\newcommand{\R}{\mathbb{R}}

\newcommand{\Expect}{\operatorname{\mathbb{E}}}

\usepackage{mathtools}

\newcommand{\Attr}{D_{\text{att}}}

\newcommand{\ImgDis}{D_{\text{adv}}}

\begin{document}
\maketitle
\begin{abstract}
We present {\em Mask-guided Generative Adversarial Network} (MagGAN) for high-resolution face attribute editing, in which semantic facial masks from a pre-trained face parser are used to guide the fine-grained image editing process. 
With the introduction of a mask-guided reconstruction loss, MagGAN learns to only edit the facial parts that are relevant to the desired attribute changes, while preserving the attribute-irrelevant regions (\eg, hat, scarf for modification `To Bald'). Further, a novel mask-guided conditioning strategy is introduced to incorporate the influence region of each attribute change into the generator. In addition, a multi-level patch-wise discriminator structure is proposed to scale our model for high-resolution ($1024 \times 1024$) face editing. Experiments on the CelebA benchmark show that the proposed method significantly outperforms prior state-of-the-art approaches in terms of both image quality and editing performance.
\end{abstract}

\vspace{-2mm}
\section{Introduction}
\label{sec:intro}
%
The demand of face editing is booming in the era of selfies. Both the research community, \eg, \cite{ChenSLLPJ_SCD,ChoiCKH0C18_StarGAN,HeZKSC17_ATTGAN,KlysSZ18_LanSubsps,LampleZUBDR17_FaderNN,LarsenSLW16,LiDH18_CRGAN,LiuDXLDZW19_STGAN,LuTT18_cCycleGAN,PerarnauWRA16_IcGAN,XieYDLT19_AMEGAN,YanYSL16_CVAE,ZhangKSC18_SaGAN,ZhangSQ17_CAAE}, and the industry, \eg, Adobe and Meitu, have extensively explored to improve the automation of face editing by leveraging user's specification of various facial attributes, \eg, hair color and eye size, as the conditional input. Generative Adversarial Networks (GANs)~\cite{goodfellow2014generative} have made tremendous progress for this task. Prominent examples in this direction include AttGAN~\cite{HeZKSC17_ATTGAN}, StarGAN~\cite{ChoiCKH0C18_StarGAN}, and STGAN~\cite{LiuDXLDZW19_STGAN}, all of which use an encoder-decoder architecture, and take both source image and target attributes (or, attributes to be changed) as input to generate a new image with the characteristic of target attributes. 

\begin{figure}[t]
    \centering
    \includegraphics[width=0.95\linewidth]{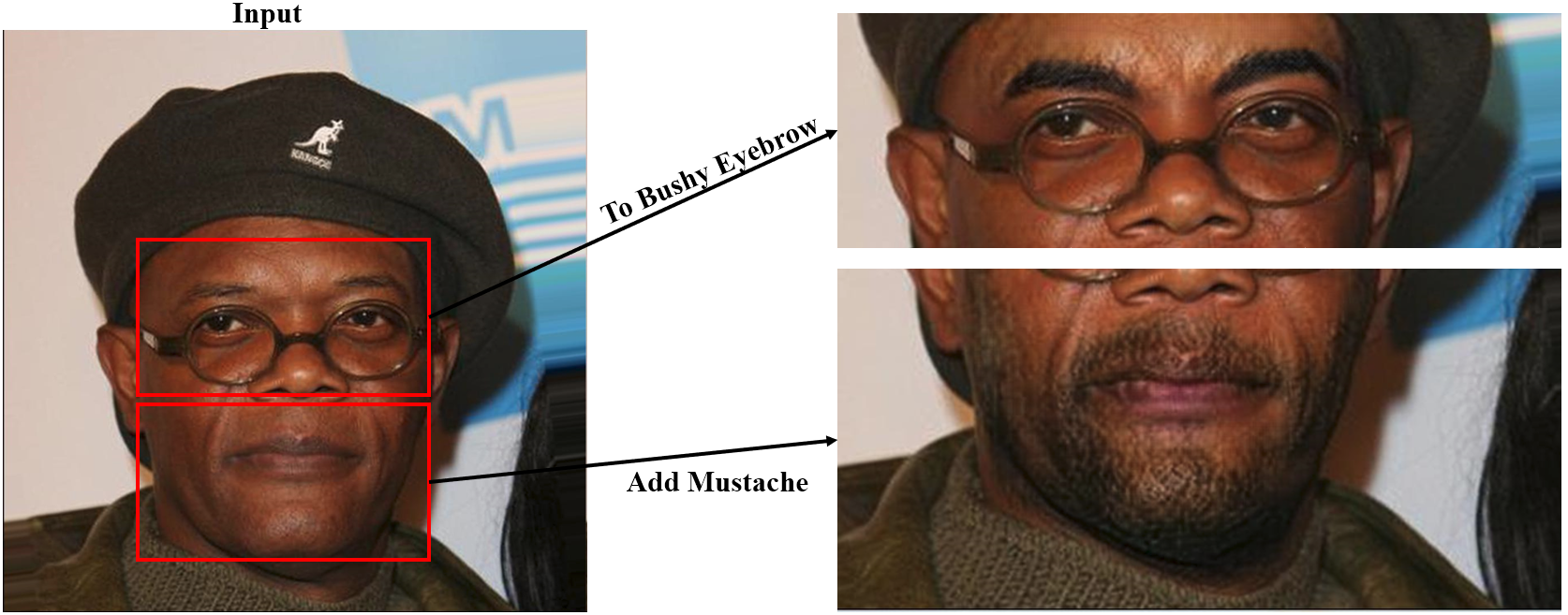}
    \includegraphics[width=0.95\linewidth]{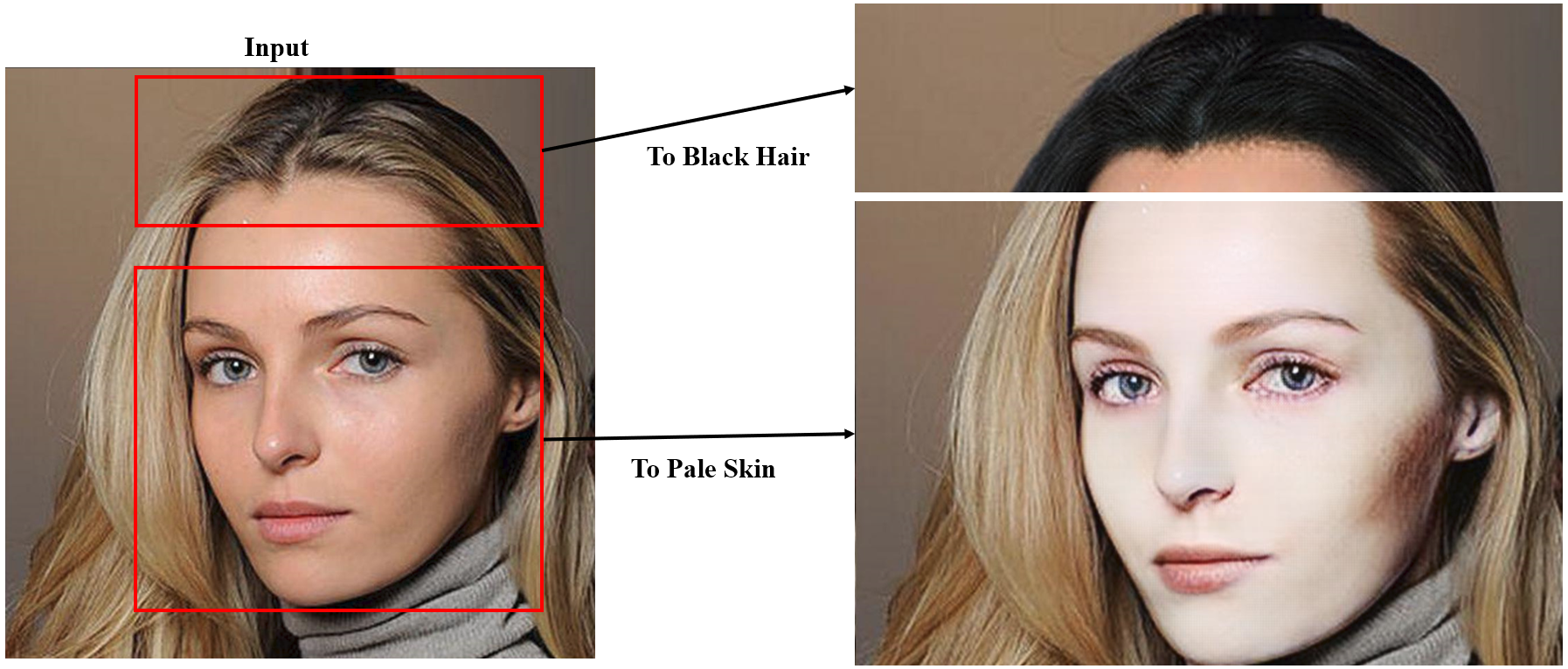}
    \label{fig:res1024}
    \vspace{-0.1cm}
    \caption{\label{fig:high_reso_results}Visual results of MagGAN on resolution $1024\times1024$. The specific sub-regions are cropped for better visualization}
    \vspace{-0.6cm}
\end{figure}

\begin{figure}[t]
\centering
\includegraphics[width=0.99\linewidth]{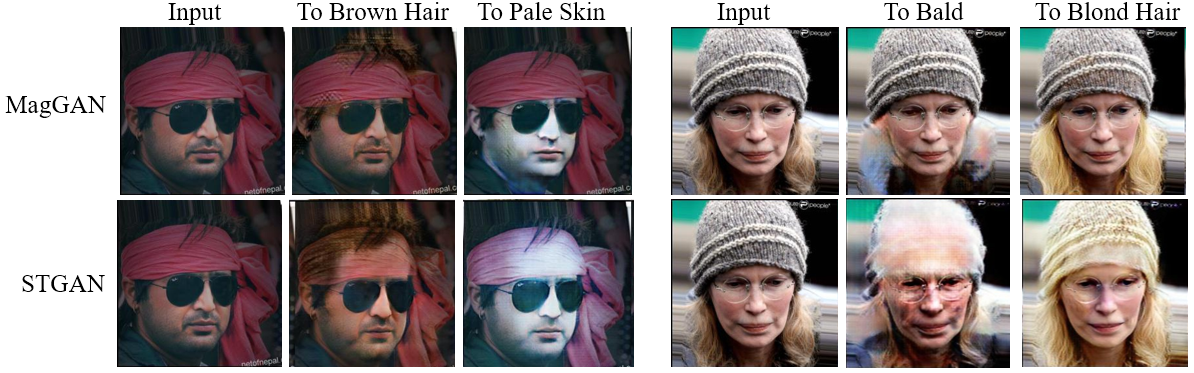}
\vspace{-0.1cm}
\caption{MagGAN (1st row) can effectively apply accurate attribute editing while keeping attribute-irrelevant regions (\emph{e.g.}, hat, scarf) intact. In comparison, the state-of-the-art STGAN~\cite{LiuDXLDZW19_STGAN} (2nd row) produces undesired modifications on these regions, \eg, whitening the scarf while manipulating ``Pale Skin'' }
\vspace{-0.2cm}
\label{fig:demo}
\end{figure}

Although promising results have been achieved, state-of-the-art methods still suffer from inaccurately localized editing, where regions irrelevant to the desired attribute change are often edited. For instance, STGAN~\cite{LiuDXLDZW19_STGAN} can make undesired editing by painting the scarf to white for ``Pale Skin'' (left) and the hat to golden for ``Blond Hair'' (right) (see Figure~\ref{fig:demo}). Solution to this problem requires notions of relevant regions that are editable {\em w.r.t.} the facial attribute edit types, while keeping the non-editable regions intact.
To illustrate this concept of region-localized attribute editing, we refer to the facial regions that are editable when a specific attribute changes as {\em attribute-relevant} regions (such as the hair region for ``To Blonde''). Regions that should not be edited (such as the hat and other non-hair regions for attribute ``To Bald'') are referred to as {\em attribute-irrelevant}. 
Ideal attribute editing generator will only edit attribute-relevant regions while keeping attribute-irrelevant regions intact, to minimize artifacts.
The second issue of most existing methods is that they only work with images of low resolutions ($128 \times 128$). How to edit facial attributes of high-resolution ($1024 \times 1024$) images is less explored. 

In order to address these challenges, we present the {\bf Mask-guided Generative Adversarial Network} (MagGAN) for high-resolution face attribute editing. The proposed approach is built upon STGAN~\cite{LiuDXLDZW19_STGAN}, which uses a difference attribute vector as conditional input, and a selective transfer unit for attribute editing. Based on this, a soft segmentation mask of common face parts from a pre-trained face parser is used to achieve fine-grained face editing. On one hand, the facial mask provides useful geometric constraints, which helps generate realistic face images. On the other hand, the mask also identifies each facial component (\eg, eyes, mouth, and hair), which is necessary for accurately localized editing. With the introduction of a mask-guided reconstruction loss, MagGAN can effectively focus on regions that are most related to the edited attributes, and keep the attribute-irrelevant regions intact, thus generating photo-realistic outputs.



Another reason why existing methods cannot preserve the regions that should not be edited is about how the attribute change information is injected into the generator. Although most attribute changes lead to localized editing, the attribute change condition itself does not explicitly contain any spatial information. In order to better learn the alignment between attribute change and regions to edit, MagGAN further uses a novel mask-guided conditioning strategy that can adaptively learn \emph{where to edit}. 



To further scale our model for high-resolution ($1024\times 1024$) face editing (see Figure~\ref{fig:high_reso_results} for visual results), we propose to use a series of multi-level patch-wise discriminators. The coarsest-level discriminator sees the full downsampled image, and is responsible for judging the global consistency of generated images, while a finer-level discriminator only sees patches of the generated high-resolution image, and tries to classify whether these patches are real or not. Empirically, this leads to more stable model training for high-resolution face editing. 

The main contributions of this paper are summarized as follows. ($i$) We propose MagGAN that can effectively leverage semantic facial mask information for fine-grained face attribute editing, via the introduction of a mask-guided reconstruction loss. ($ii$) A novel mask-guided conditioning strategy is further introduced to encourage the influenced region of each target attribute to be localized into the generator. ($iii$) A multi-level patch-wise discriminator structure scales up our model to deal with high-resolution face editing. ($iv$) State-of-the-art results are achieved on the CelebA benchmark, outperforming previous methods in terms of both visual quality and editing performance. 



\section{Related Work}
\label{sec:related}

The development of face editing techniques evolves along the automation of editing tools.
In the early stage, researchers focused on developing attribute-dedicated methods for face editing \cite{ChenCDHSSA16_InfoGAN,LiZZ16e_DIAT,LiuBK17_UNIT,ShenL17_ResImage,WangWQTL18,ZhangSXCZLQ18_SGGAN}, \ie, each model is dedicated to modifying a single attribute. However, such dedicated methods suffer from low automation level, \ie, not being able to manipulate multiple attributes in one step. To this end, many works \cite{ChoiCKH0C18_StarGAN,HeZKSC17_ATTGAN,KlysSZ18_LanSubsps,LampleZUBDR17_FaderNN,LarsenSLW16,LiDH18_CRGAN,LiuDXLDZW19_STGAN,LuTT18_cCycleGAN,PerarnauWRA16_IcGAN,XieYDLT19_AMEGAN,YanYSL16_CVAE,ZhangKSC18_SaGAN,ZhangSQ17_CAAE} started using attribute specifications, \ie, semantically meaningful attribute vectors, as conditional input. Multiple attributes can be manipulated via changing the input attribute specifications. This work belongs to this category. Another line of works \cite{ChenXTJ19_SIIT,MaJGTG18_EGIT,XiaoHM18_ELEGANT,YinLL18_GAF,ZhouXYFHH17_GeneGAN} improve the automation level of the face editing model by providing an exemplar image as the conditional input. Below, we briefly review recent attribute-specification based methods, and refer the readers to \cite{ZhengGHLH18_survey} for more details of methods that are not reviewed herein.

\begin{figure*}[t]
    \centering
    \includegraphics[width=0.99\linewidth]{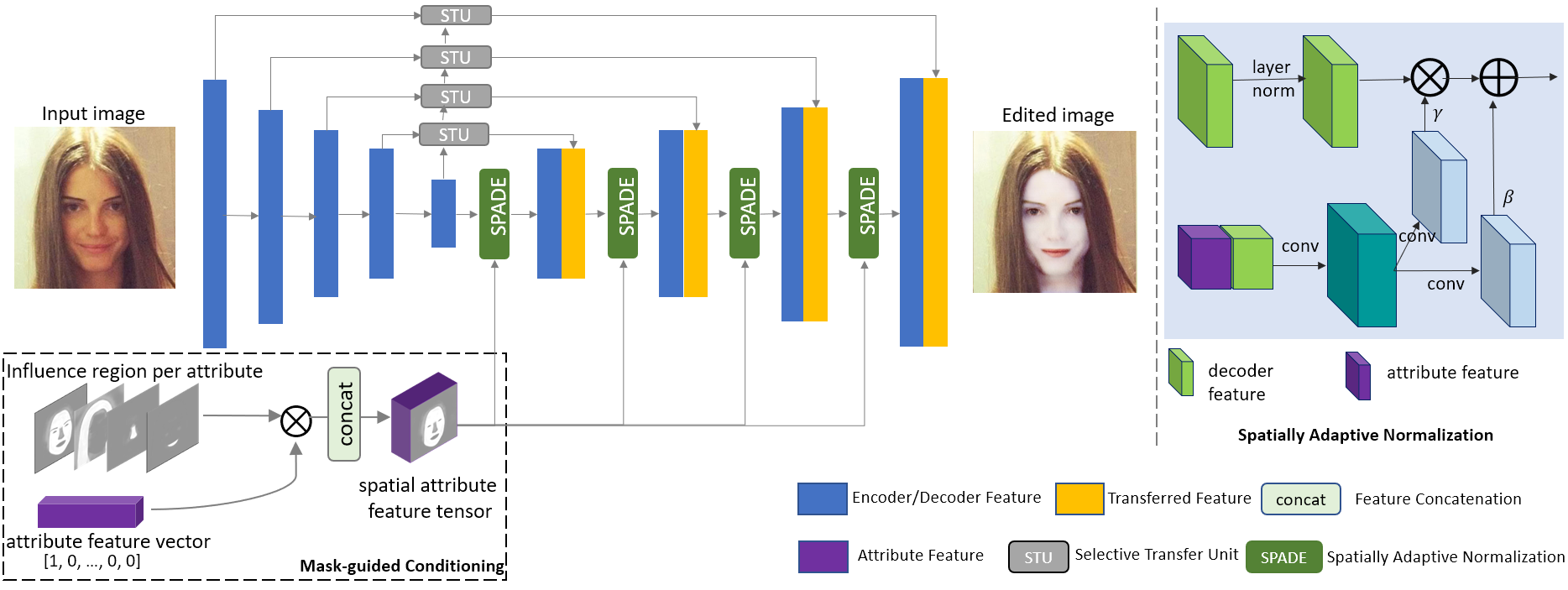}
    \vspace{-0.1cm}
    \caption{Model architecture for the proposed Mask-guided GAN (MagGAN)}
    \label{fig:architecture}
    \vspace{-0.6cm}
\end{figure*}


Many facial attributes are local properties (such as hair color, baldness, etc), and facial attribute editing should only change relevant regions and preserve regions not to be edited. StarGAN~\cite{ChoiCKH0C18_StarGAN} and CycleGAN~\cite{LuTT18_cCycleGAN} introduced the cycle-consistency loss to conditional GAN so as to preserve attribute-irrelevant details and to stabilize training. AttGAN~\cite{HeZKSC17_ATTGAN} and STGAN~\cite{LiuDXLDZW19_STGAN} found that the reconstruction loss of images not to be edited is at least as good as the cycle-consistency loss for preserving attribute-irrelevant regions. STGAN~\cite{LiuDXLDZW19_STGAN} proposed the selective transfer units to adaptively select and modify encoder features for enhanced attribute editing, achieving state-of-the-art performance on editing success rate. However, in this paper, we show that neither the cycle-consistency loss nor the reconstruction loss is sufficient to well preserve regions not to be edited (see Figure~\ref{fig:demo}), and propose to utilize masks to solve this problem.


Semantic mask/segmentation provides geometry parsing information for image generation, see, \eg, \cite{isola2017image,ParkLWZ19_SPADE,li2019object}. Semantic mask datasets and models are available for domains with important real applications, such as face editing~\cite{le2012interactive,MaskGAN} and fashion~\cite{liang2015human}. Recently, both \cite{MaskPortrait} and \cite{MaskGAN} utilize mask information for facial image manipulation, where {\it a target/manipulated mask} is required in the manipulation process. In this paper, we focus on the setting of editing with attribute specifications, without requiring a target/manipulated mask. We only make use of a pre-trained face parser, instead of requiring users to provide the mask manually.  






\vspace{-2mm}
\section{MagGAN}
\label{sec:method}
As illustrated in Figure~\ref{fig:architecture}, face editing is performed in MagGAN via an encoder-decoder architecture~\cite{HeZKSC17_ATTGAN,ChoiCKH0C18_StarGAN}. The design of Selective Transfer Units (STUs) in STGAN~\cite{LiuDXLDZW19_STGAN} is adopted to selectively transform encoder features according to the desired attribute change. Inspired by StyleGAN~\cite{KarrasLA18_StyleGAN,ParkLWZ19_SPADE}, the adaptive layer normalization~\cite{ba2016layer,huang2017arbitrary} is used to inject conditions through the de-normalization process, instead of directly concatenating the conditions with the feature map. Our full encoder-decoder generator is denoted as: 
\vspace{-2mm}
\begin{equation}
    \hat{\mathbf{x}} = G(\mathbf{x}, \mathbf{att}_{\text{diff}}), \quad \mathbf{att}_{\text{diff}} = \mathbf{att}_{t} - \mathbf{att}_{s},
\end{equation}
where $\mathbf{x} ( \text{or }\hat{\mathbf{x}}) \in \R^{3\times H \times W}$ denote the input (or edited) image; $\mathbf{att}_{s} (\text{or }\mathbf{att}_{t}) \in \R^C$ are the source (or target) attributes. The generator takes the attribute difference $\mathbf{att}_{\text{diff}} \in \R^C$ as input, following~\cite{LiuDXLDZW19_STGAN}.

\vspace{-2mm}
\subsection{Avoid editing attribute-irrelevant regions}
\label{sec:preservingmetric}

Although notable results have been achieved, existing work still suffers from inaccurately localized editing, where irrelevant regions unrelated to the desired attribute change are often made. For example, in Figure~\ref{fig:demo}, STGAN~\cite{LiuDXLDZW19_STGAN} changes the scarf to white for ``Pale Skin'' (left), and changes the hat to golden for ``Blond Hair'' (right). 

We leverage facial regions for effective facial attribute editing and modeling as a solution. We utilize a pre-trained face parser to provide soft facial region masks. Specifically, a modified BiseNet~\cite{yu2018bisenet} trained on the CelebAMask-HQ dataset~\cite{lee2019maskgan}~\footnote{\url{https://github.com/zllrunning/face-parsing.PyTorch}} is used to generates 19-class region masks, including various facial components and accessories. 
For each attribute $a_i$, we define its {\em influence regions} represented by two probability masks $M_i^{+}, M_i^{-} \in [0,1]^{H\times W}$.
If attribute $a_i$ is strengthened during editing, the region characterized by $M_i^{+}$ is likely to be changed; if $a_i$ is weakened, the region characterized by $M_i^{-}$ is likely to be changed. 
For example, for ``Pale Skin'', both $M_i^{+}$ and $M_i^{-}$ characterize the ``skin'' region; for ``Bald'', $M_i^{+}$ characterizes the ``hair'' region while $M_i^{-}$ characterizes the region consisting of ``background, skin, ears'' and ``ear rings''. In this setup, we propose the following Mask-aware Reconstruction Error (MRE) to measure the {\em preserving quality} of the editing process (in preserving irrelevant regions that shall not be edited):
\vspace{-2mm}
\small
\begin{equation}\label{eqn:MRE}
    \text{MRE} = \frac{1}{H W C} \sum_{i=1}^C \big\|(1-M_i^{\text{sgn}(\mathbf{att}_{\text{diff},i})}) (G(\mathbf{x}, \mathbf{att}_{\text{diff},i} \mathbf{e}_i) - \mathbf{x} ) 
    \big\|_1,
\end{equation}
\normalsize
where $\mathbf{att}_{\text{diff},i}$ is the $i$'th entry of $\mathbf{att}_{\text{diff}}$, and $\mathbf{e}_i$ is the vector with $i$'th entry 1 and all others 0, $M_i^{\text{sgn}(\mathbf{att}_{\text{diff},i})} \in \{ M_i^{+}$, $M_i^{-} \}$. In the face editing experiments, since all attributes are binary and $\mathbf{att}_{s} \in \{0,1\}^C$, we take the attribute change vector $\mathbf{att}_{\text{diff}} := 1 - 2 \mathbf{att}_{s}$. In this case, the image preservation error is computed when {\it only one} attribute is flipped each time, and MRE is the total error. 

In \S~\ref{sec:exp}, we will report MRE for various previous methods and our models in Table~\ref{tab:ablation}. Existing approaches of both the cycle-consistency loss used in StarGAN~\cite{ChoiCKH0C18_StarGAN} and the reconstruction loss in \cite{HeZKSC17_ATTGAN,LiuDXLDZW19_STGAN} are insufficient to preserve the regions that shall not be edited. 

\begin{figure}[t]
    \centering
    \includegraphics[width=0.9\linewidth]{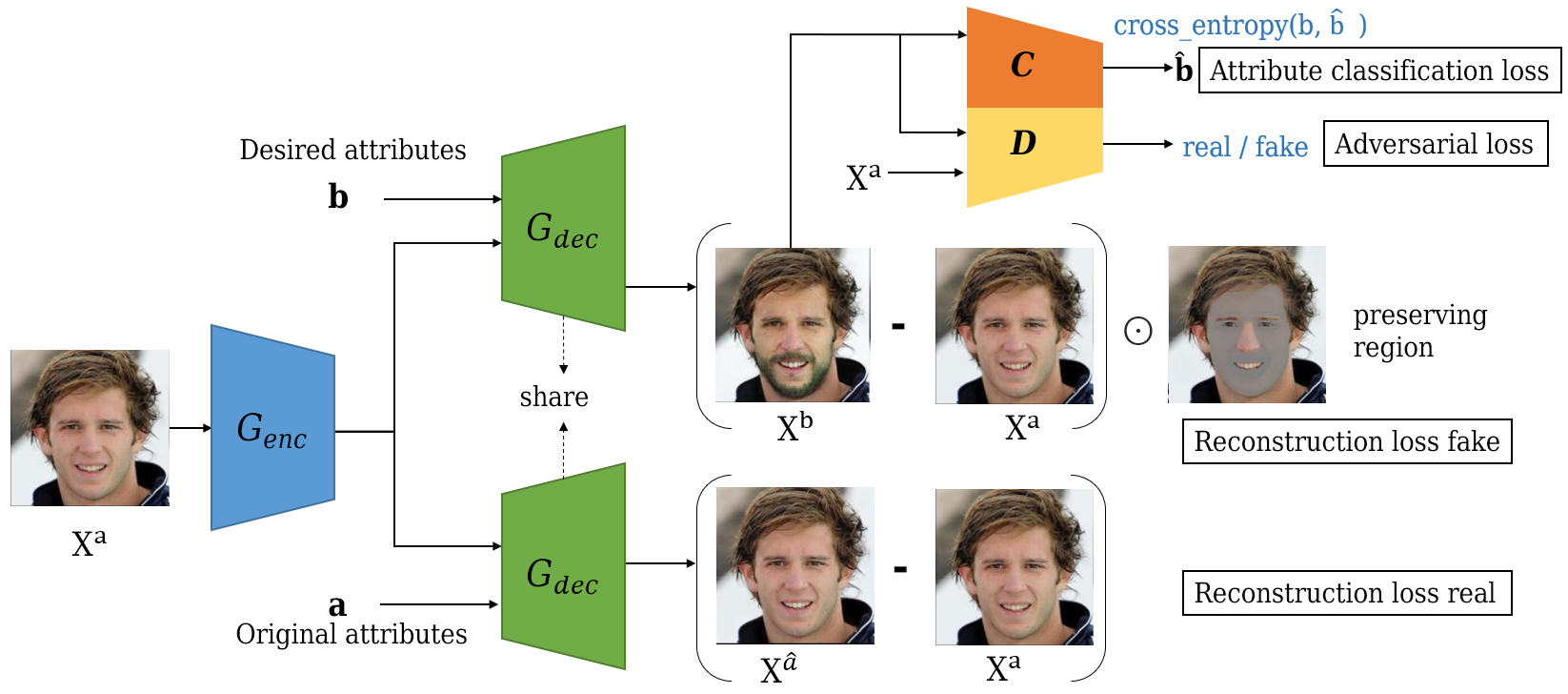}
    \vspace{-0.1cm}
    \caption{MagGAN loss function design (\S~\ref{sec:training}). For better illustration, the preserving region is denoted by the non-grey part of human face}
    \label{fig:loss}
    \vspace{-0.4cm}
\end{figure}

\vspace{-2mm}
\subsection{Loss functions for model training}
\label{sec:training}

We aim to optimize MagGAN regarding the following four aspects:
($i$) preservation accuracy for regions that should be preserved; 
($ii$) reconstruction error of the original image; 
($iii$) attribute editing success; and 
($iv$) synthesized image quality. 
Therefore, we design four respective types of loss functions for MagGAN training, as illustrated in Figure~\ref{fig:loss}.

\noindent\textbf{Mask-guided reconstruction loss.}
Continue from the design of MRE~\eqref{eqn:MRE}, we propose the following mask-guided reconstruction loss:
\vspace{-2mm}
\begin{equation}\label{eqn:MREloss}
    L_{G}^{\text{mre}} = 
    \left\|
    M(\mathbf{att}_{\text{diff}}, \mathbf{x}) \cdot (\mathbf{x} - G(\mathbf{x}, \mathbf{att}_{\text{diff}})) 
    \right\|_1,
\end{equation}
where $M(\mathbf{att}_{\text{diff}}, \mathbf{x}) \in [0,1]^{H\times W}$ is a probability mask of the regions to be preserved. 

The preserved mask $M(\mathbf{att}_{\text{diff}}, \mathbf{x})$ is computed from both the attribute difference $\mathbf{att}_{\text{diff}}$ and the probability facial mask $\mathbf{M}$ of image $\mathbf{x}$. We first feed image $\mathbf{x}$ into a face parser, and obtain a probability map $\mathbf{M} \in [0,1]^{19 \times H \times W}$ of the 19 facial parts, where $\sum^{19}_{i=1} \mathbf{M}_{i,h,w} = \mathbf{1}_{h,w}$. 
Since the semantic relationship between facial attributes and facial parts can be reasonably assumed to be constant, we explicitly define two binary relation matrices $\mathbf{AR}^{+}$ and $\mathbf{AR}^{-}$, the {\em attribute-part matrices} with dimension $C \times 19$, to characterize the relation between them. The $i$-th row of matrix $\mathbf{AR}^{+}$ or $\mathbf{AR}^{-}$ indicates which facial parts should be modified when the $i$-th attribute is strengthened, \ie, $\mathbf{att}_{\text{diff},i}>0$, or weakened, \ie, $\mathbf{att}_{\text{diff},i}<0$. Note that, if facial part has no explicit relationship with one attribute, the corresponding matrix entry of $\mathbf{AR}^{+}$,$\mathbf{AR}^{-}$ could be set to 0.

To obtain $M$, we first gather all parts $\mathbf{AR}^* \in [0,1]^{19}$ that are possibly influenced by attribute 
change $\mathbf{att}_{\text{diff}}$, as, 
\vspace{-2mm}
\begin{equation}
\mathbf{AR}^* = \min
  \left\{1, \big(\mathbf{att}_{\text{diff}}^{(+)} \big)^T \mathbf{AR}^{+} + \big(\mathbf{att}_{\text{diff}}^{(-)} \big)^T \mathbf{AR}^{-} 
  \right\},
\end{equation}
where $\mathbf{att}_{\text{diff}}^{(+)} = (\mathbf{att}_{\text{diff}}>0)$ and $\mathbf{att}_{\text{diff}}^{(-)} = (\mathbf{att}_{\text{diff}}<0)$. Finally, 
\vspace{-1mm}
\begin{equation}
M_{h,w}(\mathbf{att}_{\text{diff}}, \mathbf{x}) = \mathbf{1} - \sum_{i=1}^C \mathbf{M}_{i,h,w} * \mathbf{AR}^*_{i}.
\end{equation}

The influence regions $M_i^{+}$ and $M_i^{-}$ in \eqref{eqn:MRE} can also be computed this way, with $\mathbf{att}_{\text{diff}} = \mathbf{e}_i$ and $\mathbf{att}_{\text{diff}} = -\mathbf{e}_i$. 

\noindent\textbf{Reconstruction loss.} 
Image reconstruction can be considered as a sub-task of image editing, because the generator should reconstruct the image when no edit is applied, $\mathbf{att}_{\text{diff}} = \mathbf{0}$. Therefore, the reconstruction loss is defined as
\vspace{-2mm}
\begin{equation}\label{eq:loss_recon}
    \mathcal{L}_{G}^{\text{rec}} = \|G(\mathbf{x}, \mathbf{0}) - x\|_1,
\end{equation}
where the $\ell_1$ norm is adopted to preserve the sharpness of the reconstructed image. 

\noindent\textbf{GAN loss for enhancing image quality.}
The synthesized image quality is enhanced by the generative adversarial networks, where we use an unconditional image discriminator $\ImgDis$ to differentiate real images from edited images. In particular, a Wasserstein GAN (WGAN)~\cite{arjovsky2017wasserstein} is utilized:
\vspace{-2mm}
\begin{equation}\label{eq:loss_gan_dis}
\begin{aligned}
    \mathcal{L}_{\ImgDis} = & \Expect_{\hat{\mathbf{x}}}[\ImgDis(\hat{\mathbf{x}})] - \Expect_{\mathbf{x}}[\ImgDis(\mathbf{x})] +  \lambda \Expect_{\mathbf{x}_{\text{int}}}[(\| \nabla_{\mathbf{x}_{\text{int}}}\ImgDis(\mathbf{x}_{\text{int}}) \|_2 - 1)^2],
\end{aligned}
\end{equation}
where $\hat{\mathbf{x}}$ is the generated image and $\mathbf{x}_{\text{int}}$ is sampled along lines between the latent space of pairs of real and generated image.  

The generator $G$, instead, tries to fool the discriminator by synthesizing more realistic images:
\vspace{-2mm}
\begin{equation}\label{eq:loss_gan}
\begin{aligned}
   \mathcal{L}_{G}^{\text{gan}} =  - \Expect_{\mathbf{x}, \mathbf{att}_{\text{diff}}}[\ImgDis(G(\mathbf{x}, \mathbf{att}_{\text{diff}}))].
\end{aligned}
\end{equation}
\vspace{-2mm}

\noindent\textbf{Attribute classification loss.} 
To ensure that the edited image indeed has the target attribute $\mathbf{att}_t$, an attribute classifier $\Attr$ is trained on the ground-truth image attribute pairs $(\mathbf{x}, \mathbf{att}_s)$ with the standard cross-entropy loss:
\vspace{-2mm}
\begin{equation}
    \mathcal{L}_{\Attr} = \Expect_{\mathbf{x}}[ KL(\Attr(\mathbf{x}), \mathbf{att}_s) ]\,.
\end{equation}
The generator is trying to generate images that maximize its probability to be classified with the target attribute $\mathbf{att}_t$:
\vspace{-2mm}
\begin{equation}\label{eq:loss_attrcls}
    \mathcal{L}_{G}^{\text{cls}} = - \Expect_{\mathbf{x}, \mathbf{att}_{\text{diff}}}[ KL(\Attr(G(\mathbf{x}, \mathbf{att}_{\text{diff}})), \mathbf{att}_t) ]\,.
\end{equation}
In summary, the loss to train the MagGAN generator $G$ is 
\vspace{-2mm}
\begin{equation}
    \mathcal{L}_{G} = {L}_{G}^{\text{gan}} + \lambda_1 \mathcal{L}_{G}^{\text{rec}} + \lambda_2 \mathcal{L}_{G}^{\text{cls}} + \lambda_3 L_{G}^{\text{mre}}.
\end{equation}
In experiments, we always take $\lambda_1=100$ and $\lambda_2 = 10$. We vary $\lambda_3$ to examine the effect of our proposed mask-guided reconstruction loss.  

\vspace{-2mm}
\subsection{Mask-guided conditioning in the generator}
\label{sec:maskgenerator}


Another reason why the previous methods cannot preserve the regions that shall not be edited is about how the attribute change information is injected into the generator. Although most attribute changes should lead to localized editing, the attribute change condition $\mathbf{att}_{\text{diff}} \in \R^{C}$ does not explicitly contain any spatial information. In STGAN~\cite{LiuDXLDZW19_STGAN} (and other previous works for face attribute editing), this condition is replicated to have the same spatial size of some hidden feature tensor, and then concatenated to it in the generator. For example, in the SPADE block in Figure~\ref{fig:generator} (Right), $\mathbf{att}_{\text{diff}}$ is replicated spatially to be $\mathbf{Att}_{\text{diff}} \in \R^{C\times H \times W}$ (the purple block)\footnote{We use $\mathbf{att} \in \R^{C}$ to denote attributes without spatial dimension and $\mathbf{Att} \in \R^{C\times H\times W}$ for attributes with spatial dimensions.}, and then concatenated to the decoder feature (the green block). It is hoped that the generator will learn by itself the localized property of attribute editing from this concatenated tensor. However, in practice, this is insufficient, even with the mask-guided reconstruction loss~\eqref{eqn:MREloss}. 

We propose to inject this inductive bias that the influence region of each attribute change is localized into the generator directly, by making use of masks. We view the $i$-th channel of $\mathbf{Att}_{\text{diff}}$, denoted as $\mathbf{Att}_{\text{diff}}^{(i)} \in \R^{H\times W}$, as the condition to edit attribute $a_i$. In previous work, $\mathbf{Att}_{\text{diff}}^{(i)} = \mathbf{att}_{\text{diff},i} \mathbf{1}$ that is uniform across the spatial dimension. Specifically, we propose:
\vspace{-2mm}
\begin{equation}
    \mathbf{Att}_{\text{diff}}^{(i)} = \mathbf{att}_{\text{diff},i} M_i^{\text{sgn}(\mathbf{att}_{\text{diff},i})},
\end{equation}
where $M_i^{+}$ and $M_i^{-}$ are the influence regions of attribute $a_i$ defined in \eqref{eqn:MRE}. We illustrate this mask-guided conditioning process in Figure~\ref{fig:generator} (bottom-left). Finally, we simply replace the original replicated tensor with the mask-guided attribute condition tensor, and obtain {\it a generator with mask-guided conditioning}. 
Note that this mask-guided conditioning technique is generally applicable to both generators with and without SPADE. 

\noindent\textbf{The blending trick} is another simple approach to preserve the attribute-irrelevant regions. More specifically, with the probability mask of attribute-irrelevant regions $M(\mathbf{att}_{\text{diff}}, \mathbf{x})$ defined in \eqref{eqn:MREloss}, we simply add a linear layer at the end of the generator:
\vspace{-2mm}
\begin{equation}
    \hat{x} = M(\mathbf{att}_{\text{diff}}, \mathbf{x})*x + 
    \left( 1 - M(\mathbf{att}_{\text{diff}}, \mathbf{x}) 
    \right)*G(\mathbf{x}, \mathbf{att}_{\text{diff}}).
\end{equation}
This blending trick improves our MagGAN performance in terms of MRE, but visually it introduces sharp transitions at the boundary of regions to be preserved. Therefore, we do not include this trick in our final MagGAN. More discussions are in Appendix~\ref{sec:res256}.

\vspace{-2mm}
\subsection{Multi-level patch-wise discriminators for high-resolution face editing}
\label{sec:patchdiscriminator}

\begin{figure}[t]
    \centering
    \includegraphics[width=0.85\linewidth]{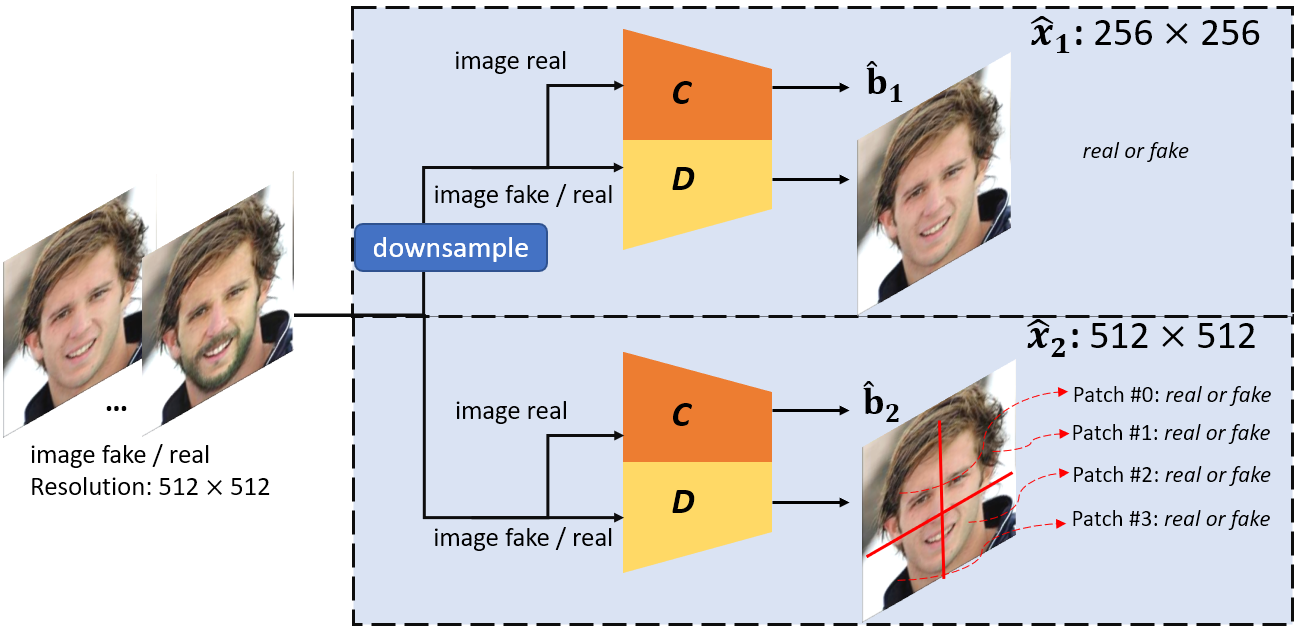}
    \vspace{-0.3cm}
    \caption{Illustration of multi-level patch-wise discriminators}
    \label{fig:patchdis}
    \vspace{-0.6cm}
\end{figure}

We describe our approach to scale up image editing in high resolutions. First of all, we empirically found that a single ``shallow'' discriminator cannot learn some global concepts, such as Male/Female, leading to low editing success. On the other hand, a single ``deep'' discriminator makes the adversarial training very unstable, leading to low image quality.

Inspired by PatchGAN~\cite{isola2017image} and several multi-level generation works~\cite{Han16stackgan,Han17stackgan2,karras2017progressive}, we propose to use a series of multi-level patch-wise ``shallow'' discriminators, as illustrated in Figure~\ref{fig:patchdis}, for high-resolution face editing. The architecture of the discriminators are exactly the same without sharing weights. The coarsest-level discriminator ($D_1$) see the full downsampled image, and is responsible for global consistency in the image generation. The attribute classifier $C_1$ associated with it is effective in attribute classification, as in the low-resolution image editing case. The finer-level discriminators ($D_2$, etc.) see patches of the generated high-resolution image instead of the full one, and determine whether these patches are real or not. To maintain an unified architecture for discriminators across different levels, we still associate the finer-level discriminator with a classifier ($C_2$), which takes the average pooled feature as input for classification. The total loss for all PatchGAN discriminators are defined as:
\vspace{-2mm}
\begin{equation}
    \mathcal{L}_{D} = \frac{1}{P}\sum^{P}_{i=1} \big( \mathcal{L}_{\Attr^i} + \mathcal{L}_{\ImgDis^i} \big),
\end{equation}
where $\Attr^i$, $\ImgDis^i$ denote the attribute classifier and image discriminator of the $i$th PatchGAN discriminator, $P$ is the number of total discriminators. In practice, we found these finer-level discriminators improve the editing performance.

Note that our generator only generates high-resolution images, which can be directly downsampled to lower resolutions and fed to coarse-level discriminators. 
On the contrary, generators in previous works~\cite{Han16stackgan,Han17stackgan2,karras2017progressive} generate a high-resolution image in a multi-stage manner for the sake of training stability. They generate low-resolution images as intermediate outputs, which are fed to coarse-level discriminators. Our approach is simple in comparison, and we did not observe any training stability issue. 

\vspace{-2mm}
\section{Experiments}
\label{sec:exp}

\textbf{Dataset and pre-processing.} We use CelebA dataset \cite{LiuLWT15} for evaluation. CelebA contains over 200K facial images with 40 binary attribute labels for each image. To apply CelebA to high-resolution face editing, we process the original web images by cropping, aligning and resizing into $1024\times1024$. When loading images for editing, they are re-scaled to match the target resolution. The images are divided into the training set, validation set and test set. Following the repository of STGAN\footnote{STGAN: \url{https://github.com/csmliu/STGAN}}, we take 637 images from the validation set to assess the training process. We use the rest of the validation set and the training set to train our model. The test set (nearly 20K) is used for evaluation. We consider 13 distinctive attributes including: \emph{Bald}, \emph{Bangs}, \emph{Black Hair}, \emph{Blond Hair}, \emph{Brown Hair}, \emph{Bushy Eyebrows}, \emph{Eyeglasses}, \emph{Male}, \emph{Mouth Slightly Open}, \emph{Mustache}, \emph{No Beard}, \emph{Pale Skin} and \emph{Young}. Since most images in CelebA have lower resolution than $1024\times 1024$, our ``high-resolution" MagGAN models are not exactly trained with true high-resolution images. However, our results show the ability of MagGAN scale up to $1024\times 1024$ resolution.

MagGAN exploits the information of facial masks, which are obtained using a pre-trained face parser with 19 classes (as mentioned in \S~\ref{sec:preservingmetric}). Instead of taking a multi-label hard mask, we take the probability of each class as soft masks with smooth boundaries, which leads to improved generation quality. All the facial masks are stored in resolution $256\times256$. The two attribute-part relation matrices $\text{AR}^+, \text{AR}^- \in [0,1]^{13\times19}$ described in \S~\ref{sec:training} characterize the relation between each edit attribute and corresponding facial component changes. Detailed definitions are in Appendix~\ref{sec:ar_matrix}. 

\noindent\textbf{Quantitative evaluation.} The performance of attribute editing are measured in three aspects, \ie, (\emph{i}) mask-aware reconstruction error (MRE),  (\emph{ii}) attribute editing accuracy and (\emph{iii}) image quality. 

\begin{figure*}[t]
    \centering
    \includegraphics[width=1\linewidth]{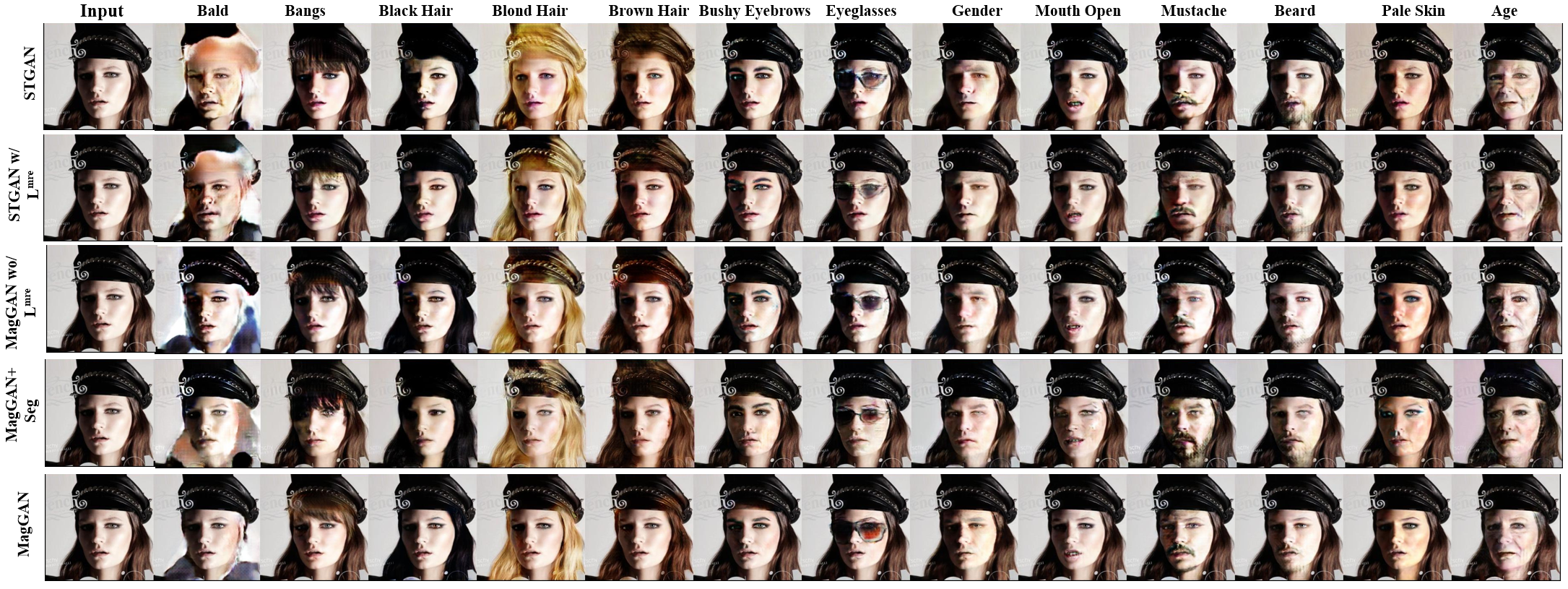}
    \vspace{-0.3cm}
    \caption{Visual results of MagGAN variants on resolution $256\times256$. Each column represents edited images through one attribute reversing editing} 
    \label{fig:ablation}
    \vspace{-0.6cm}
\end{figure*}

Table~\ref{tab:transfer} shows that MagGAN decreases the MRE significantly, indicating better preserving of regions that should be intact. This improvement is also obvious in the editing results in Figure~\ref{fig:ablation}. Table~\ref{tab:transfer} also reports the PSNR/SSIM score of the reconstructed image by keeping target attribute vector the same as the source one (definition in Appendix~\ref{sec:metric}). MagGAN also improves PSNR/SSIM significantly.

\begin{figure}[t]
    \centering
    \includegraphics[width=0.9\linewidth]{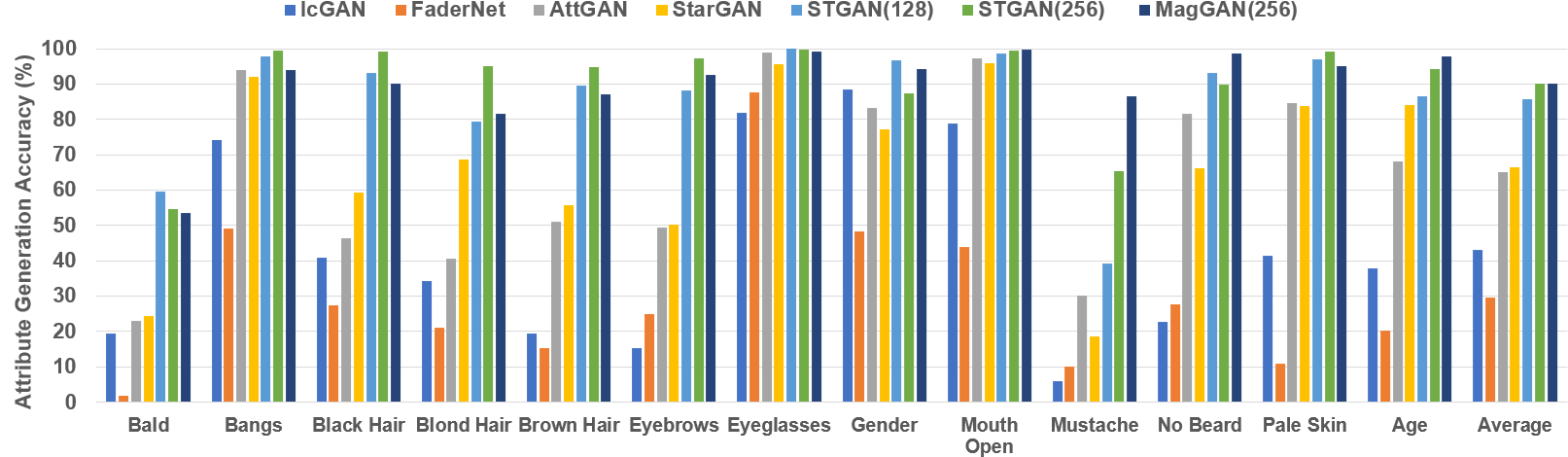}
    \vspace{-0.2cm}
    \caption{Facial attribute editing accuracy of IcGAN \cite{PerarnauWRA16_IcGAN}, FaderNet \cite{LampleZUBDR17_FaderNN}, StarGAN \cite{ChoiCKH0C18_StarGAN}, AttGAN \cite{HeZKSC17_ATTGAN}, STGAN  \cite{LiuDXLDZW19_STGAN}, STGAN(256) and our model MagGAN(256) (from left to right in rainbow colors in order). The last two models naming with ``(256)" are the ones with image resolution 256 that are resized into 128 for evaluation}
    \label{fig:acc}
    \vspace{-0.6cm}
\end{figure}

\begin{figure}[t]
    \centering
    \includegraphics[width=0.92\linewidth]{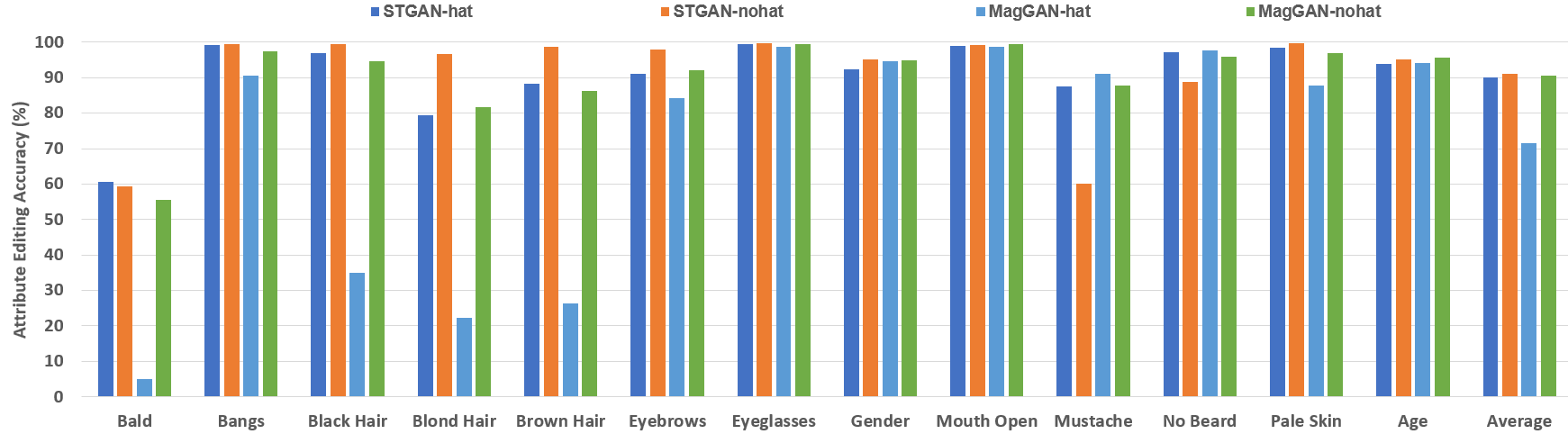}
    \vspace{-0.2cm}
    \caption{Facial attribute editing accuracy of STGAN and MagGAN on hat samples and non hat samples of resolution $256\times256$}
    \label{fig:acc_hat}
    \vspace{-0.6cm}
\end{figure}

\begin{table}[t]
\caption{Comparison of quantitative results with SOTA}
\label{tab:transfer}
\centerline{
\begin{tabular}{|l||c||c||c||c||c|}
\hline
Methods & \makebox[3.5em]{MRE $\downarrow$}  & \makebox[3.5em]{FID $\downarrow$} & \makebox[3.5em]{Avg Acc} & \makebox[3.5em]{PSNR} & \makebox[3.5em]{SSIM} \\ \hline \hline
AttGAN(128) & $ 0.0713 $ & $ 10.23 $ & $ 64.9\% $ & $24.07 $ & $0.841 $ \\ \hline
STGAN(128) & 0.0627 & $ 7.75 $ & $ 85.8\% $ & $31.67 $ & $0.948 $\\ \hline
STGAN(256) & 0.0530 & $ 1.21 $ & $ {90.4}\% $ & $ 37.61 $ & $ 0.959 $   \\ \hline
MagGAN(256) & {0.0163} & $ \textbf{1.10} $ & $ {90.0}\% $ & $ 40.25 $ & $0.984$ \\ \hline
MagGAN(512) & 0.0141 & $ 1.20 $ & $ {89.1}\% $ & $ 41.42 $ & $0.987 $ \\ \hline
MagGAN(1024) & \textbf{0.0130} & $ 1.31 $ & $ \textbf{91.0}\% $ & $ \textbf{42.94} $ & $\textbf{0.994} $ \\ \hline
\end{tabular}
}
\vspace{-0.6cm}
\end{table}

We also report the attribute editing accuracy by employing the pre-trained attribute classification model from \cite{LiuDXLDZW19_STGAN}.
We follow the evaluation protocol used in \cite{HeZKSC17_ATTGAN,LiuDXLDZW19_STGAN}. For each test image, reverse one of its 13 attributes at a time ($1 \rightarrow 0$ or $0 \rightarrow 1$), and generate an image after each reversion; so there are 13 edited images for each input image. The widely used evaluation metric is \emph{attribute editing accuracy}, which measures the successful manipulation rate for the reversed attribute each time, but ignores the attribute preservation error. Figure~\ref{fig:acc} reports the facial attribute manipulation accuracy of previous works IcGAN \cite{PerarnauWRA16_IcGAN}, FaderNet \cite{LampleZUBDR17_FaderNN}, AttGAN \cite{HeZKSC17_ATTGAN}, StarGAN \cite{ChoiCKH0C18_StarGAN}, STGAN \cite{LiuDXLDZW19_STGAN} and our proposed MagGAN. To build the strongest baseline, we also train our own STGAN model at resolution $256\times 256$, optimizing all possible parameters; see details of the hyperparameter tuning in Appendix~\ref{sec:arch}. 

\noindent\textbf{High editing accuracy \vs~attribute-irrelevant region preserving.} As shown in Table~\ref{tab:transfer}, MagGAN at resolution 256 outperforms all the previous reported numbers except STGAN(256) on average accuracy. In Figure~\ref{fig:acc}, compared with STGAN(256), MagGAN(256) is better in "Mustache", "No beard", "Gender", "Age" and worse in "Bald", "Bangs", "Black Hair", "Blonde Hair", "Brown Hair". We conjecture that STGAN(256) achieves this high accuracy by editing hat or scarf when they appear in the image; like coloring the hat to golden to get an editing success of "Blonde Hair". To verify this assumption, we separate the testing set into two groups -- samples with hat, samples without hat by measuring the area ratio of hat in the face masks (we select threshold 0.1 to decide if the sample contains a hat). The attribute editing accuracy is evaluated on the two subsets respectively. Results in Figure~\ref{fig:acc_hat} show that the editing accuracy of MagGAN decreases a lot on hat subset on several hat-related attributes, \eg, ``Bald'', ``Black Hair'', but on par with STGAN on non hat subsets. In this sense, MagGAN editing success is even higher than our strongest baseline STGAN(256) since it can preserve the attribute irrelevant regions, making editing more real. 

To measure the image quality, we report FID (Fréchet Inception Distance) score \cite{heusel2017FID}. The FID score measures the distance between the Inception-v3\footnote{We pretrained an Inception-V3 model that achieves 92.69\% average attribute classification accuracy on all 40 attributes of CelebA dataset.} activation distributions of original images and the edited images. 
Table~\ref{tab:transfer} shows that the FID score improves significantly from resolution 128 to 256, but then get stalled and insensitive to image quality for 256 and higher resolutions. This is because the input size for Inception-v3 model is 299, and thus resolution increase from 128 to 256 is significant. However, all high resolution generations are first downsampled to evaluate the FID score. After all, MagGAN at all resolutions achieves the comparable result with the best FID score. Finally, due to smaller batches in training for high resolutions, FID scores of MagGAN(512) and MagGAN(1024) are slightly lower than those of MagGAN(256).

\noindent\textbf{Qualitative evaluation.} Apart from the quantitative evaluation, we visualize some facial attribute editing results at resolution $256\times256$ in Figure~\ref{fig:ablation}, and compare our proposed model with the state-of-the-art method, \ie, STGAN \cite{LiuDXLDZW19_STGAN} (as it is the strongest baseline) and other variations.

\begin{table}[t]
\caption{Results of user study for ranking methods on two subsets considering hat wearing}
\label{tab:user_study}
\centerline{
\begin{tabular}{|l|c|c|c|}
\hline
Winner method & \makebox[3.5em]{w/ hat} & \makebox[3.5em]{w/o hat} & \makebox[3.5em]{Overall} \\ \hline
MagGAN &  \textbf{59.2}\% &  \textbf{52.1}\% & \textbf{55.7}\% \\ \hline
STGAN  &  37.7\% &  45.3\% & 41.5\% \\ \hline 
Tie    &  3.1 \% &  2.6\%  & 2.8 \% \\ \hline
\end{tabular}
}
\vspace{-0.5cm}
\end{table}

\noindent\textbf{User Study.} We conduct user study on Amazon Turk to compare the generation quality of STGAN and MagGAN. To verify that MagGAN performs better on editing attribute relevant regions, we randomly choose 100 input samples from test set, 50 samples with hat or scarf and 50 samples without (since STGAN usually fails on person wearing hat). For each sample, 5 attribute editing tasks are performed by STGAN and MagGAN (500 comparison pairs in total). All 5 tasks are randomly chosen from 13 attributes, for subjects with hat, we increase the chance to select hair related attributes. The users are instructed to choose the best result which changes the attribute more successfully considering image quality and identity preservation. To avoid human bias, each sample pair is evaluated by 3 volunteers. The results are shown in Table~\ref{tab:user_study}, MagGAN outperforms STGAN on both hat samples and without hat samples. 

\vspace{-2mm}
\section{Ablation Study}
\label{sec:ablation}

We conduct three groups of ablation comparisons in image resolution of $256\times256$, to verify the effectiveness of the proposed modules individually: ($i$) mask guided reconstruction loss, ($ii$) spatially modified attribute feature, and ($iii$) usage of SPADE normalization.

We consider seven variants, \ie, ($i$) STGAN: STGAN at resolution $256\times256$, ($ii$) STGAN+cycle: STGAN with cycle-consistency loss instead of its original reconstruction loss, ($iii$) STGAN w/ $L^{\text{mre}}$: STGAN plus mask guided reconstruction loss, ($iv$) MagGAN w/o $L^{\text{mre}}$: MagGAN trained without mask guided reconstruction loss, ($v$) MagGAN w/o SP:  MagGAN without using SPADE, ($vi$) MagGAN: our proposed model with the usage of mask-guided reconstruction loss and make-guided attribute conditioning. ($vii$) MagGAN+Seg: Instead of using a pre-trained face parser, build a face segmentation branch (adopting FCN\cite{long2015fcn} architecture) into generator as sub-task, making the whole model fully trainable.


\noindent\textbf{Mask-guided reconstruction loss.}
We compare three reconstruction loss: (i) STGAN with only the reconstruction loss computed by reconstructed images, (ii) cycle-consistency loss which is applied in StarGAN \cite{ChoiCKH0C18_StarGAN}, (iii) two parts of reconstruction loss (computed on reconstructed images and synthesized images respectively) proposed in $\S$~\ref{sec:training}.
Row 1-3 of Table \ref{tab:ablation} report the quantitative results of STGAN applying each type of reconstruction loss respectively. We observe that adding mask guided reconstruction loss to generator training can effectively reduce Mask-aware Reconstruction Error (MRE). In Figure~\ref{fig:ablation}, the synthesized image of STGAN w/ $L^{\text{mre}}$ on attribute ``Bald'' and ``Blonde Hair'' also proves this assumption. But since the spatial information of mask is not directly injected into generator, STGAN w/ $L^{\text{mre}}$ still cannot preserve the attribute-irrelevant regions well.

\begin{table}[t]
\caption{Comparison of variants of MagGAN on $256\times 256$}
\label{tab:ablation}
\centerline{
\resizebox{0.95\textwidth}{15mm}{
\begin{tabular}{|l||c||c||c||c|c|}
\hline
Methods & \makebox[5em]{MRE $\downarrow$} & \makebox[5em]{FID $\downarrow$} & \makebox[5em]{Avg Acc} & \makebox[5em]{PSNR} & \makebox[5em]{SSIM}  \\ \hline \hline
(i) STGAN & 0.0530 & $ 1.21 $ & $ 90.4\% $ & $ 37.61 $ & $ 0.959 $   \\ \hline
(ii) STGAN+cycle & 0.0530 & $ 1.31 $ & $ 87.3\% $ & $ 36.14 $ & $ 0.970 $   \\ \hline
(iii) STGAN w/ $L^{\text{mre}}$ & 0.0289 & $ 1.33 $ & \textbf{95.6}\% & $ 38.48 $ & \textbf{0.984}  \\ \hline
(iv) MagGAN w/o $L^{\text{mre}}$ & 0.0397 & $1.22$ & $89.6\% $ & $ 39.35 $ & $ 0.980 $ \\ \hline
(v) MagGAN w/o SP & \textbf{0.0161} & $1.23$ & $ 89.9\% $ & \textbf{40.40} & $ 0.982 $ \\ \hline
(vi) MagGAN & 0.0163& \textbf{1.10}  & $ 90.0\% $ & $ 40.25 $ & \textbf{0.984} \\ \hline
(vii) MagGAN+Seg & 0.0612 & 2.39 & 90.3\% & 40.10 & 0.983 \\ \hline
\end{tabular}
}}
\vspace{-0.6cm}
\end{table}

\noindent\textbf{Mask-guided attribute conditioning.}
Utilizing mask-guided attribute conditioning instead of the spatially uniformed attribute conditioning provides generator with more spatial information of the interest regions. From Table~\ref{tab:ablation}, (i) \vs (iv), (iii) \vs (vi) illustrate that the MRE score decreases obviously when mask-guided attribute conditioning is applied in generator. It implies that generator effectively takes the regions of interest and edits on these local regions. 
Taking advantage of both mask-guided reconstruction loss and attribute conditioning strategy, MagGAN achieves the best MRE and FID. And the visual results in Figure~\ref{fig:ablation} also show that MagGAN makes accurate editing on hair related attributes ('Bold', 'Blonde Hair', \etc), by preserving the region of hat while only remove or paint the hair. 
MagGAN w/o SP and MagGAN perform nearly the same as (v) \vs (vi), which demonstrates the denormalization method does not affect much on performance.
Finally, the quantitative results and visual results of (vii) MagGAN+Seg are bad, which indicates the incorporating mask segmentation branch as part of the generator is not a good choice. Since the mask-guided reconstruction loss and attribute conditioning requires accurate masks, training segmentation branch with generator from scratch makes the model hard to train and undermines the editing accuracy.

\begin{figure}[t]
    \centering
    \includegraphics[width=0.92\linewidth]{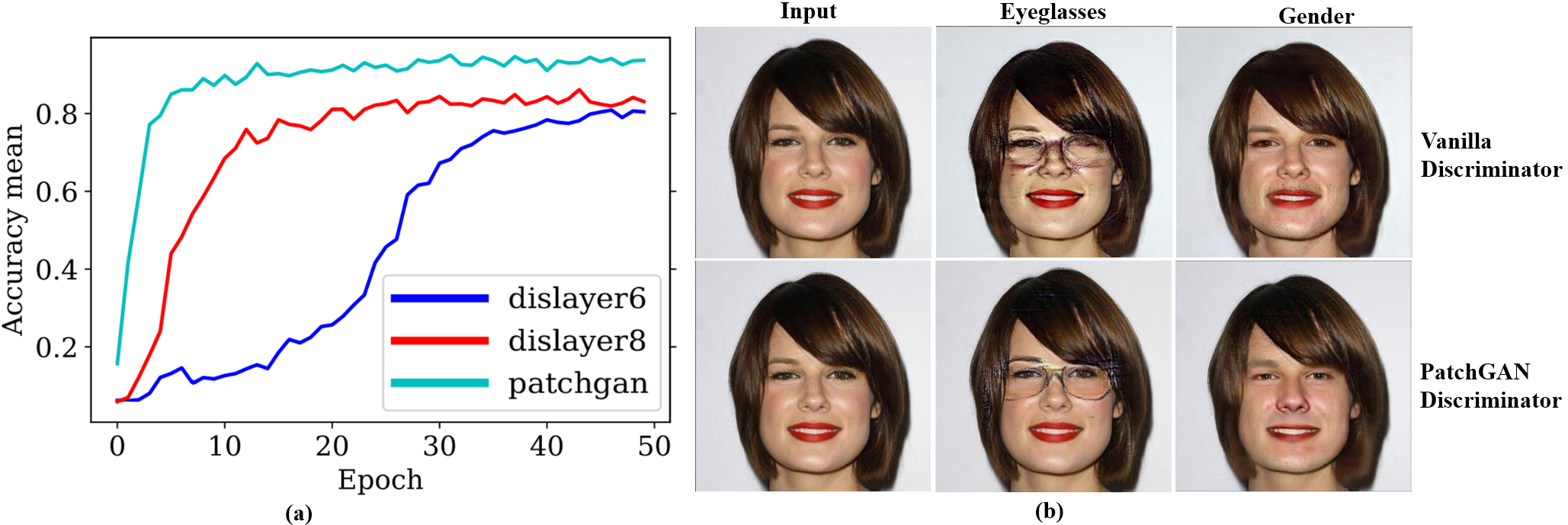}
    \vspace{-0.3cm}
    \caption{Comparison of training with vanilla single discriminator and multi-level PatchGAN discriminators on resolution $1024\times1024$: (a) attribute editing accuracy and (b) visual results}
    \label{fig:patchgan_comp}
    \vspace{-0.6cm}
\end{figure}

\vspace{5pt}
\noindent\textbf{Multi-level PatchGAN discriminator for high resolution editing.}
We apply PatchGAN discriminator to supervise training of high resolution image generation. We are able to scale the generated image resolution up to $1024\times1024$. 
In Figure~\ref{fig:patchgan_comp}, we compare the $1024\time1024$ version of training with a single discriminator and with our proposed multi-level PatchGAN discriminators. Under this setting, PatchGAN has 3 discriminators working on resolution $256\times256$, $512\times512$ and $1024\time1024$, respectively. In Figure~\ref{fig:patchgan_comp} (a), when applying single vanilla discriminator, the generator converges slower than using PatchGAN discriminator and early stops at low editing accuracy. In Figure~\ref{fig:patchgan_comp} (b), editing effects on ``Eyeglasses'', ``Gender'' from PatchGAN are more obvious than original discriminator. We assume PatchGAN discriminators provide more supervise signal on global and local regions, thus helping generator learns more discriminative features for each attribute. See more visual results in Appendix.


\vspace{-2mm}
\section{Conclusion}
In this paper, we propose MagGAN for high-resolution face image editing. The key novelty of our work lies in the use of facial masks for achieving more accurate local editing. Specifically, the mask information is used to construct a mask-guided reconstruction loss and mask-guided conditioning in the generator. MagGAN is further scaled up for high-resolution face editing with the help of PatchGAN discriminators. To our knowledge, it is the first time face attribute editing is able to be applied on resolution $1024\times1024$.

\clearpage








\makeatletter
\newcommand{\@chapapp}{\relax}%
\makeatother

\appendix
\section*{Appendix}
\renewcommand{\thesubsection}{\Alph{subsection}}

This appendix has 5 sections. \S~\ref{sec:arch} describes the network architecture of MagGAN. \S~\ref{sec:res256} shows some facial editing results on resolution $256\times256$. \S~\ref{sec:res_high} demonstrates some visual results on high resolution $512\times512$ and $1024\times1024$. \S~\ref{sec:ar_matrix} defines the attribute and facial part relationship matrix  $\mathbf{AR}^{+}$,  $\mathbf{AR}^{-}$. \S~\ref{sec:metric} provides formal definition of evaluation metrics - PSNR and SSIM.

\subsection{Network Architecture of MagGAN}
\label{sec:arch}
We present MagGAN network architecture for image generators in Table~\ref{tab:gen_struct} and the network architectures for discriminators in Table~\ref{tab:dis_strut}. They are built with basic blocks defined in Table~\ref{tab:basic_blocks}. We reached this architecture design by extensive architecture search based on STGAN, as we present below.

\begin{figure}[t]
    \centering
    \includegraphics[width=0.95\linewidth]{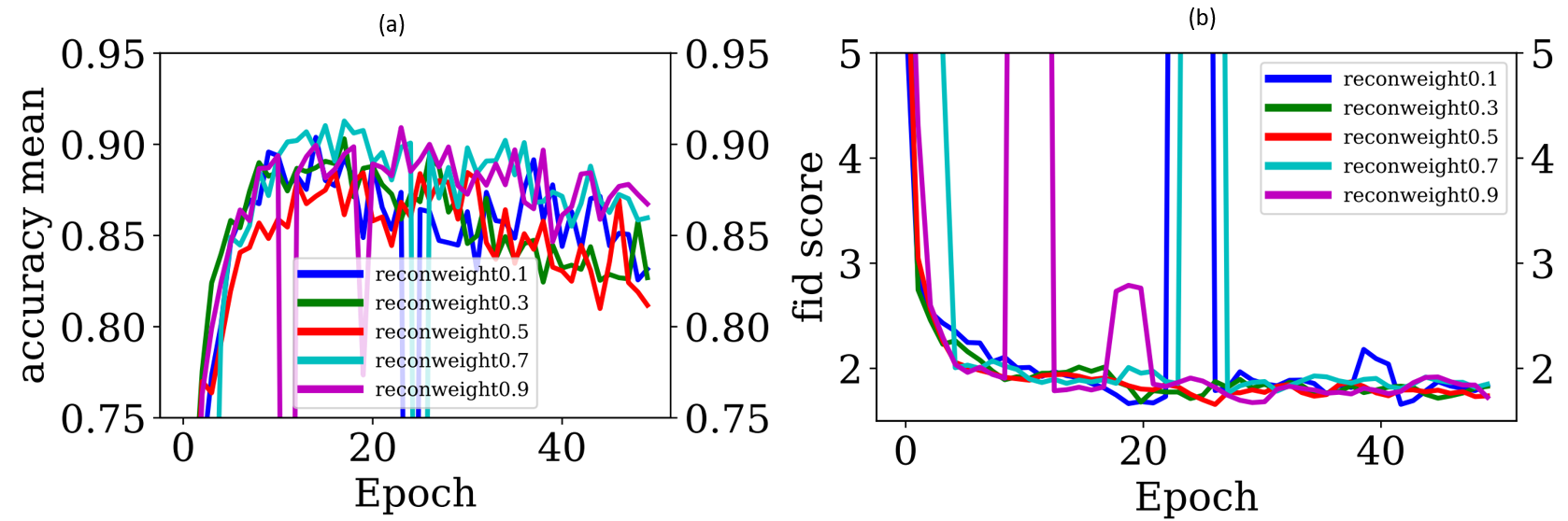}
    \caption{Attribute editing accuracy and FID comparison for reconstruction weight tuning. 'reconweight' ranges from 0 to 1, 'reconweight' = 0 means that only cycle-consistency loss is applied, 'reconweight' = 1 means that only reconstruction loss is applied}
    \label{fig:reconweight}
\end{figure}

\begin{figure}[t]
    \centering
    \includegraphics[width=0.95\linewidth]{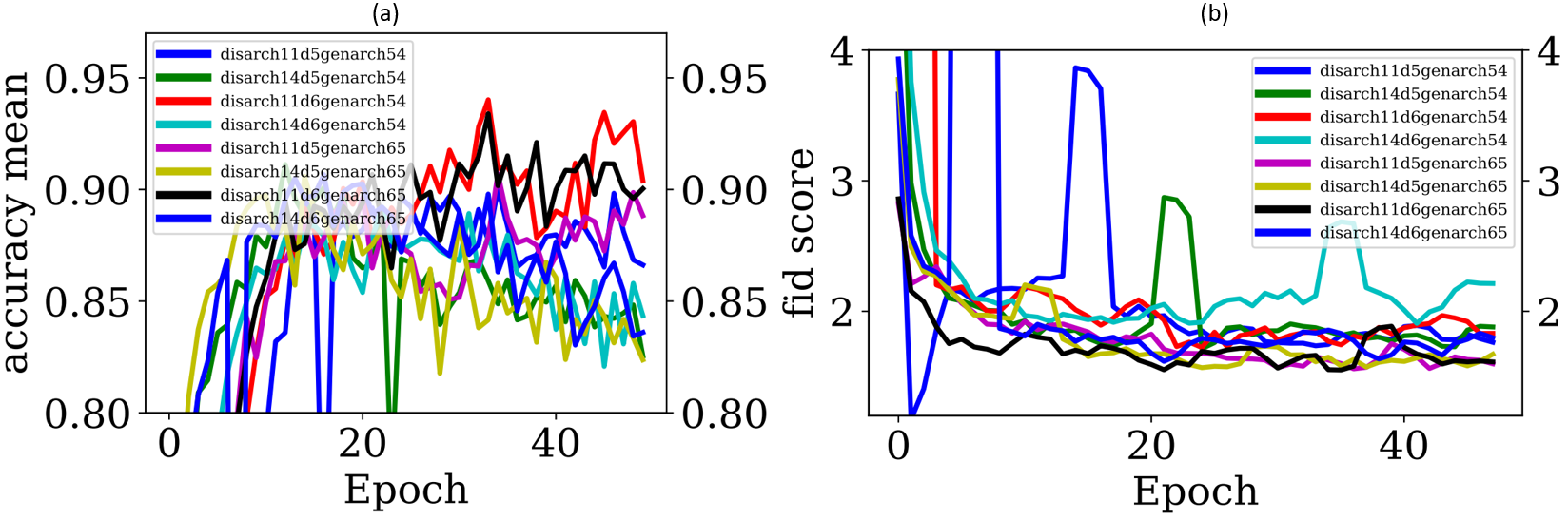}
     \caption{Attribute editing accuracy and FID comparison for architecture search. 'disarch11' means that average pooling is applied after the last shared convolution layer, output size is 1. 'disarch14' means no average pooling after the final convolution layer. 'd5', 'd6' means that the discriminator has 5 or 6 shared convolution layers. 'genarch54', 'genarch65' means that generator has 5 encoding-decoding layers, 4 STU layers or 6 encoding-decoding layers, 5 STU layers}
    \label{fig:arch_search}
\end{figure}

\paragraph{Architecture optimization based on STGAN} We first conduct hyper-parameter tuning for STGAN~\cite{LiuDXLDZW19_STGAN} on resolution $256\times256$, and compare the attribute editing accuracy and FID to select the best architecture. First, we apply both cycle-consistency loss \cite{ChoiCKH0C18_StarGAN} $\mathcal{L}^{cycle}$ and the reconstruction loss used in AttGAN~\cite{HeZKSC17_ATTGAN} $\mathcal{L}^{rec}$ to train generator, but combine the two losses with a weight $C \in [0,1]$. Then the total reconstruction loss $\mathcal{L}^{R}_G$ is defined as: $$\mathcal{L}^{R}_G = C\cdot \mathcal{L}^{rec} + (1-C)\cdot \mathcal{L}^{cycle}$$
From Figure~\ref{fig:reconweight}, we find that only applying the reconstruction loss achieves the best accuracy and FID. From Figure~\ref{fig:arch_search}, we also find that increasing the layer of discriminator and generator from 5 to 6 improves the attribute editing accuracy and FID. Also, in the original STGAN discriminator, images are fed into a shared convolution layer, and the feature maps are then used by two separate branches for adversarial prediction and attribute classification. We observe that applying average pooling after the shared convolution layer improves the attribute editing accuracy. 

\begin{figure}[t!]
    \centering
    \includegraphics[width=0.99\linewidth]{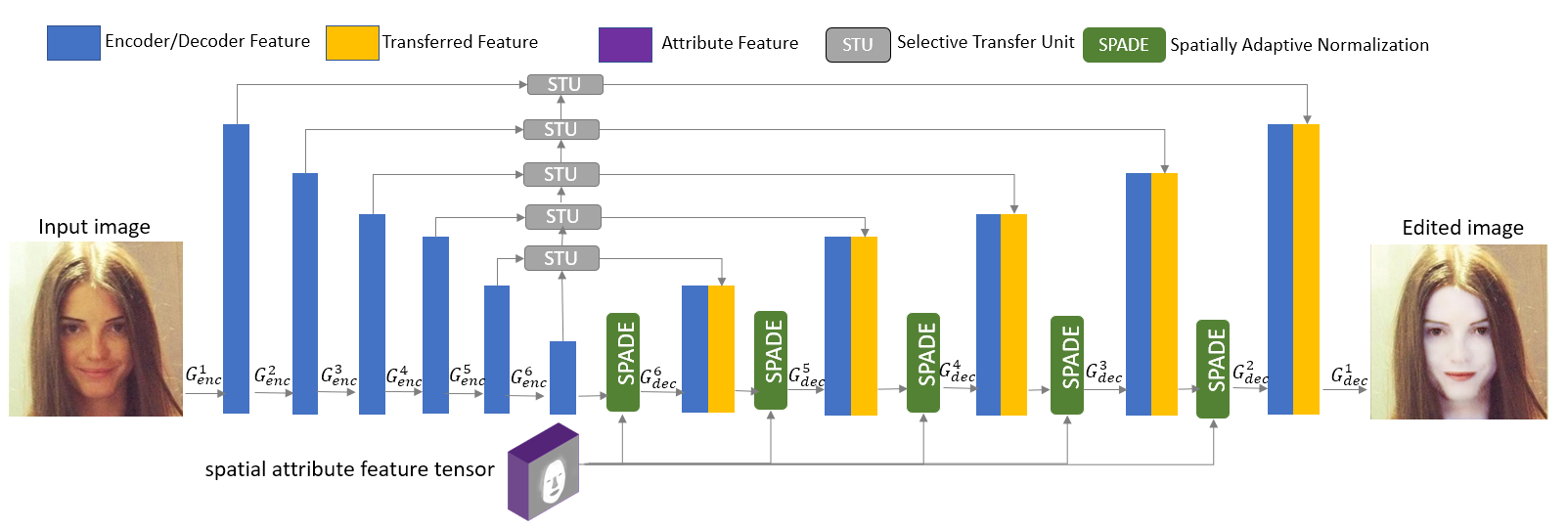}
    \caption{Network architecture of MagGAN generator}
    \label{fig:generator}
\end{figure}

\begin{figure}[t!]
    \centering
    \includegraphics[width=0.85\linewidth]{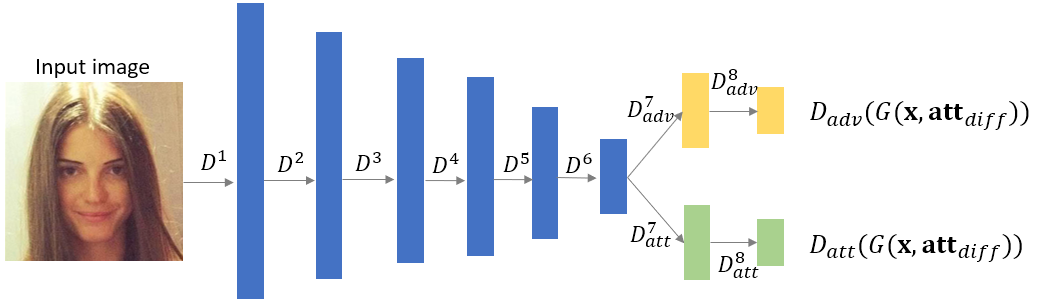}
    \caption{Network architecture of MagGAN discriminator}
    \label{fig:discriminator}
\end{figure}

\begin{table}[t]
\centering
\caption{The basic blocks for architecture design. (``-" connects two consecutive layers; ``+" means element-wise addition between two layers, * means element-wise multiplication between two layers.) $F_{dec}$, $F_{attr}$, $F_{trans}$ denotes decoding feature, spatial attribute feature, transferred feature as shown in Figure~\ref{fig:generator}}
\begin{tabular}[t]{|p{2.7cm}|p{9.3cm}|}\hline
{Name} &{\makebox[10cm]{Operations / Layers}}\\ \hline\hline
Concat & Concatenate input tensors along the channel dimension.  \\ \hline
{Downsample} & Nearest neighbor Downsampling layer \\ \hline
{BN} & Batch normalization layer \\ \hline
{IN} & Instance normalization layer \\ \hline
{LN wo/ affine} & Layer normalization layer without apply affine transformation \\ \hline
{Conv($dim$, $k$, $s$)} & Convolutional layer with output dimension $dim$, kernel size $k$, stride $s$ \\ \hline
{DeConv($dim$, $k$, $s$)} & Transposed convolutional layer with output dimension $dim$, kernel size $k$, stride $s$ \\ \hline
{STU} & Selective transfer unit proposed by STGAN~\cite{LiuDXLDZW19_STGAN} \\ \hline
\multirow{ 4}{*}{SPADE} & Spatially adaptive normalization layer~\cite{ParkLWZ19_SPADE}: \\
& $\beta$ = Concat($F_{dec}$, Downsample($F_{attr}$)) - Conv($d$,$3$,$1$) - Conv($d$,$3$,$1$)  \\
& $\gamma$ = Concat($F_{dec}$, Downsample($F_{attr}$)) - Conv($d$,$3$,$1$) - Conv($d$,$3$,$1$) \\
& LN wo/ affine ($F_{dec}$) - Conv($d$, $3$, $1$) * $\gamma$ + $\beta$  \\ \hline
Avgpool ($os$) & Average pooling with output size $os$. \\\hline
\end{tabular}
\label{tab:basic_blocks}
\vspace{-0.3cm}
\end{table}

\begin{table}[t]
    \centering
    \caption{Network architecture of MagGAN generator. $G_{enc}^{l}$ and $G_{dec}^{l}$ denotes the encoding layer and decoding layer of generator at layer $l$ respectively. The input feature of DeConv layer is the concatenation of decoding feature and selective feature. SPADE is applied as normalization layer for transposed convolution feature}
    \begin{tabular}{|c|c|c|} \hline
        $l$ & $G_{enc}^{l}$ & $G_{dec}^{l}$  \\ \hline
        1 & Conv(64, 4, 2), BN, Leaky ReLU & DeConv(3, 4, 2), Tanh \\ \hline
        2 & Conv(128, 4, 2), BN, Leaky ReLU & DeConv(128, 4, 2), SPADE, ReLU \\ \hline
        3 & Conv(256, 4, 2), BN, Leaky ReLU & DeConv(256, 4, 2), SPADE, ReLU \\ \hline
        4 & Conv(512, 4, 2), BN, Leaky ReLU & DeConv(512, 4, 2), SPADE, ReLU \\ \hline
        5 & Conv(1024, 4, 2), BN, Leaky ReLU & DeConv(1024, 4, 2), SPADE, ReLU \\ \hline
        6 & Conv(1024, 4, 2), BN, Leaky ReLU & DeConv(1024, 4, 2), SPADE, ReLU \\ \hline
    \end{tabular}
    \label{tab:gen_struct}
    \vspace{-0.3cm}
\end{table}

\begin{table}[t]
\caption{Network architecture of MagGAN discriminator/PatchGAN discriminator. $i$ denote the level of PatchGAN discriminator, $i= \{0, 1, 2\}$ corresponds to resolution 256, 512, 1024 respectively. When $i=0$, PatchGAN discriminator is equal to single vanilla discriminator applied on resolution 256. $c$ denotes the attribute class numbers. $D_{adv}^{l}$ and $D_{att}^{l}$ denotes the adversarial learning branch and attribute classification branch respectively, they share the same convolution backbone}
\centering
\begin{tabular}{|c|c|c|} \hline
    $l$ & $D_{adv}^{l}$ & $D_{att}^{l}$ \\ \hline
    1 & \multicolumn{2}{c|}{Conv($64$, $4$, $2$), IN, Leaky ReLU} \\ \hline
    2 & \multicolumn{2}{c|}{Conv($128$, $4$, $2$), IN, Leaky ReLU} \\ \hline
    3 & \multicolumn{2}{c|}{Conv($256$, $4$, $2$), IN, Leaky ReLU} \\ \hline
    4 & \multicolumn{2}{c|}{Conv($1024$, $4$, $2$), IN, Leaky ReLU} \\ \hline
    5 & \multicolumn{2}{c|}{Conv($1024$, $4$, $2$), IN, Leaky ReLU} \\ \hline
    6 & \multicolumn{2}{c|}{Conv($1024$, $4$, $2$), IN, Leaky ReLU} \\ \hline
    7 &  \multicolumn{2}{c|}{Avgpool($2^i$)} \\ \hline
    8 & Conv(1024, 1, 1), Leaky ReLU & Conv(1024, 1, 1), Leaky ReLU \\ \hline
    9 & Conv(1, 1, 1) & Conv(c, 1, 1) \\ \hline
\end{tabular}
\label{tab:dis_strut}
\end{table}

\begin{figure}[t]
    \centering
    \includegraphics[width=0.95\linewidth]{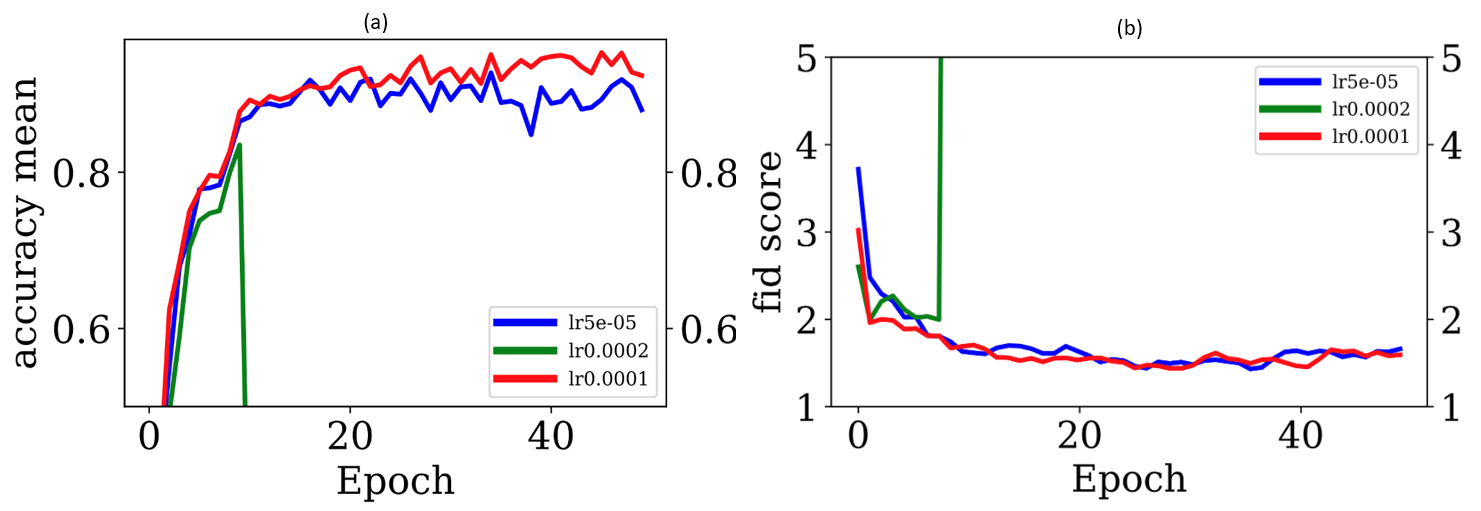}
    \caption{Attribute editing accuracy and FID comparison for generator learning rate tuning. We test 3 learning rate: $5\times10^{-5}$, $1\times10^{-4}$, $2\times10^{-4}$}
    \label{fig:lr_tune}
\end{figure}

\paragraph{Learning rate optimization to stablize training} In our experiment setting, we set encoding/decoding layer of generator to 6. The shared convolution backbone layer of the vanilla discriminator or PatchGAN discriminators is also set to 6. The illustration of network architecture for generator and discriminators are shown in Figure~\ref{fig:generator} and Figure~\ref{fig:discriminator}. To make generator training stable, the learning rate of generator is set to $0.0001$ according to Figure~\ref{fig:lr_tune}, while learning rate of discriminator is set to $0.0002$.

\paragraph{Hyper-parameter for mask-guided reconstruction weight} With extensive experiments, we find that to achieve reasonable visual effects for synthesized images, both mask-guided attribute conditioning and mask-guided reconstruction loss should be applied. We present the effects of mask-guided reconstruction weight $\lambda_3$ in Table~\ref{tab:l4}. To achieve a better balance between editing accuracy and preserving quality, we choose $\lambda_3=200$ in practice.


\paragraph{Unified architecture for a single discriminator and multi-level patch-wise discriminators} We unify the architecture of vanilla STGAN discriminator and PatchGAN discriminator in Table~\ref{tab:dis_strut}. The difference is that level-$i$ PatchGAN discriminator works on different resolution, from 256 to 1024, the adversarial outputs of PatchGAN discriminators ($i=0,1,2$) are of size $(1,1)$, $(2,2)$, $(4,4)$ respectively. Each output entry represents a real/fake output corresponding to a $256\times256$ patch.

\clearpage
\subsection{Face Attribute Editing Results on Resolution 256}
\label{sec:res256}
\begin{figure}[t]
    \centering
    \includegraphics[width=0.99\linewidth]{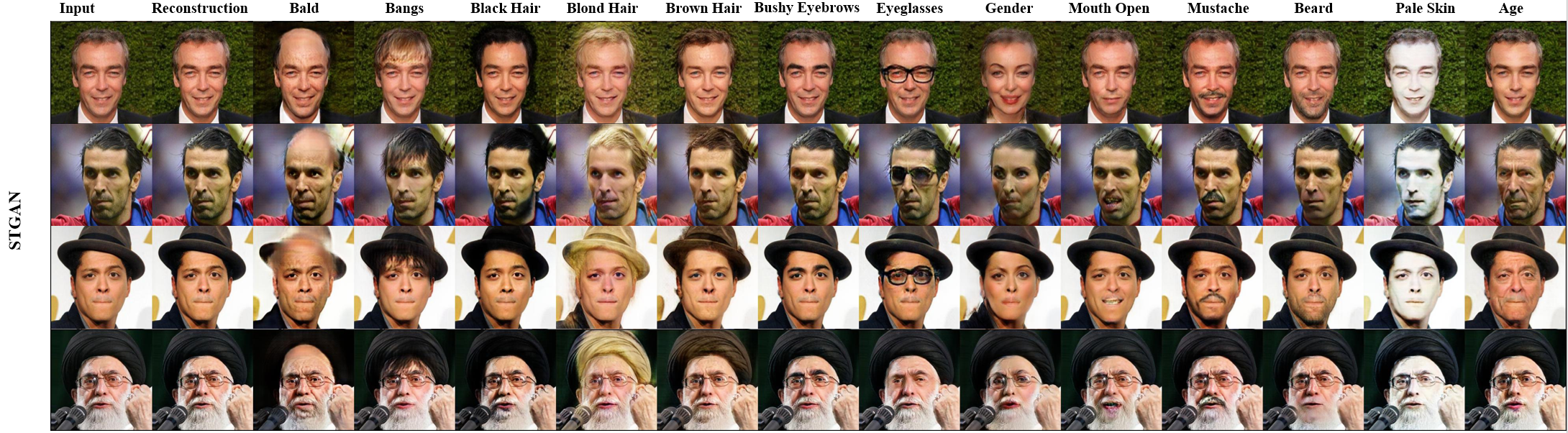}
    \includegraphics[width=0.99\linewidth]{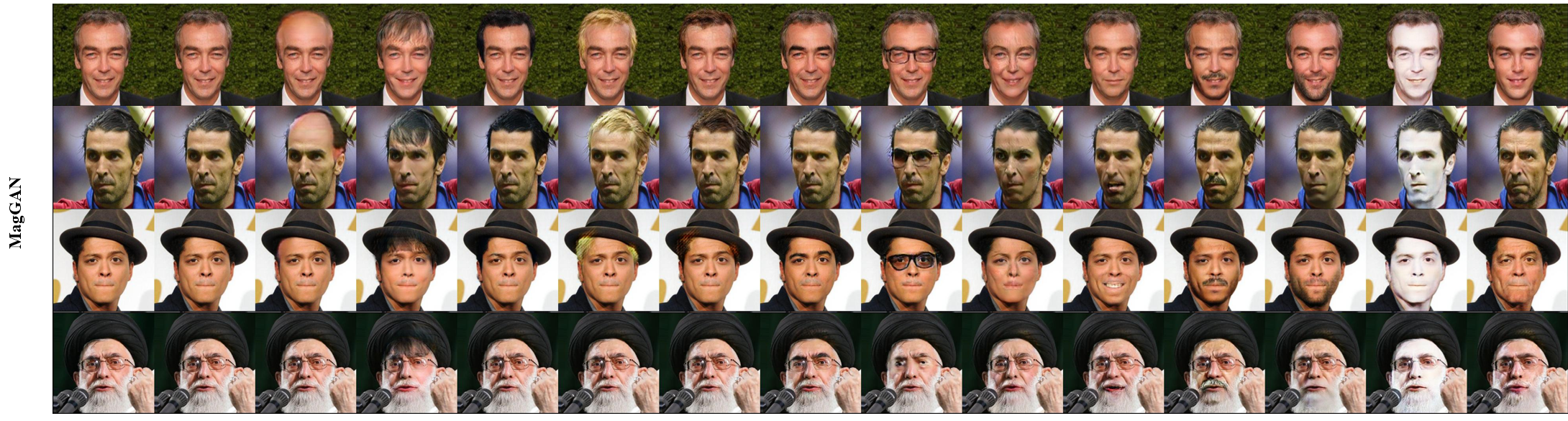}
    \includegraphics[width=0.99\linewidth]{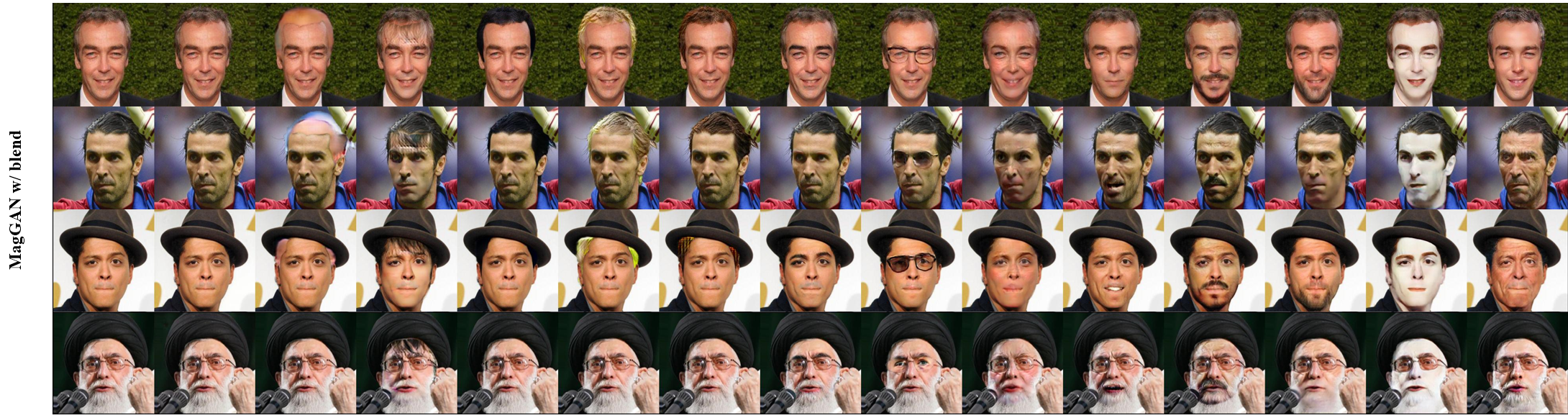}
    \caption{Visual results of STGAN, MagGAN, MagGAN w/ blend on resolution $256\times256$}
    \label{fig:maggan_res}
\end{figure}

\begin{figure}
    \centering
    \includegraphics[width=0.9\linewidth]{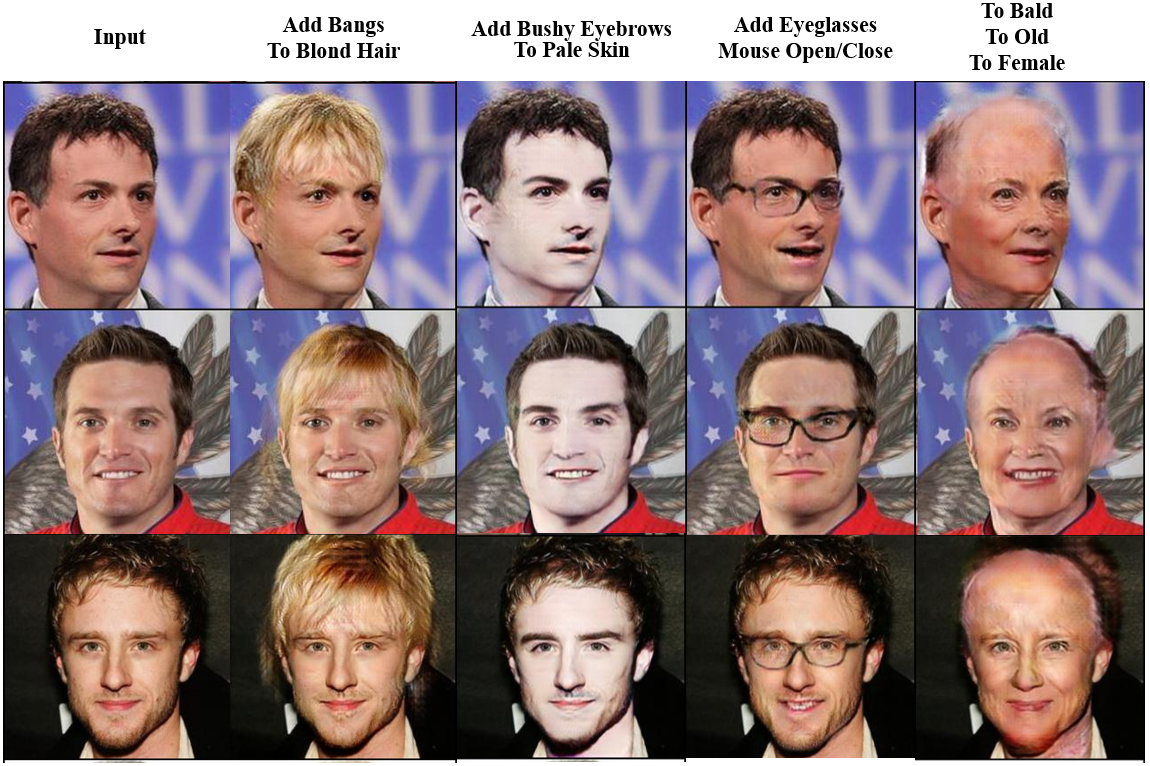}
    \caption{Visual results of MagGAN for multiple facial attribute editing on resolution $256\times256$}
    \label{fig:maggan_multi}
\end{figure}

\begin{figure}[t]
    \centering
    \includegraphics[width=0.99\linewidth]{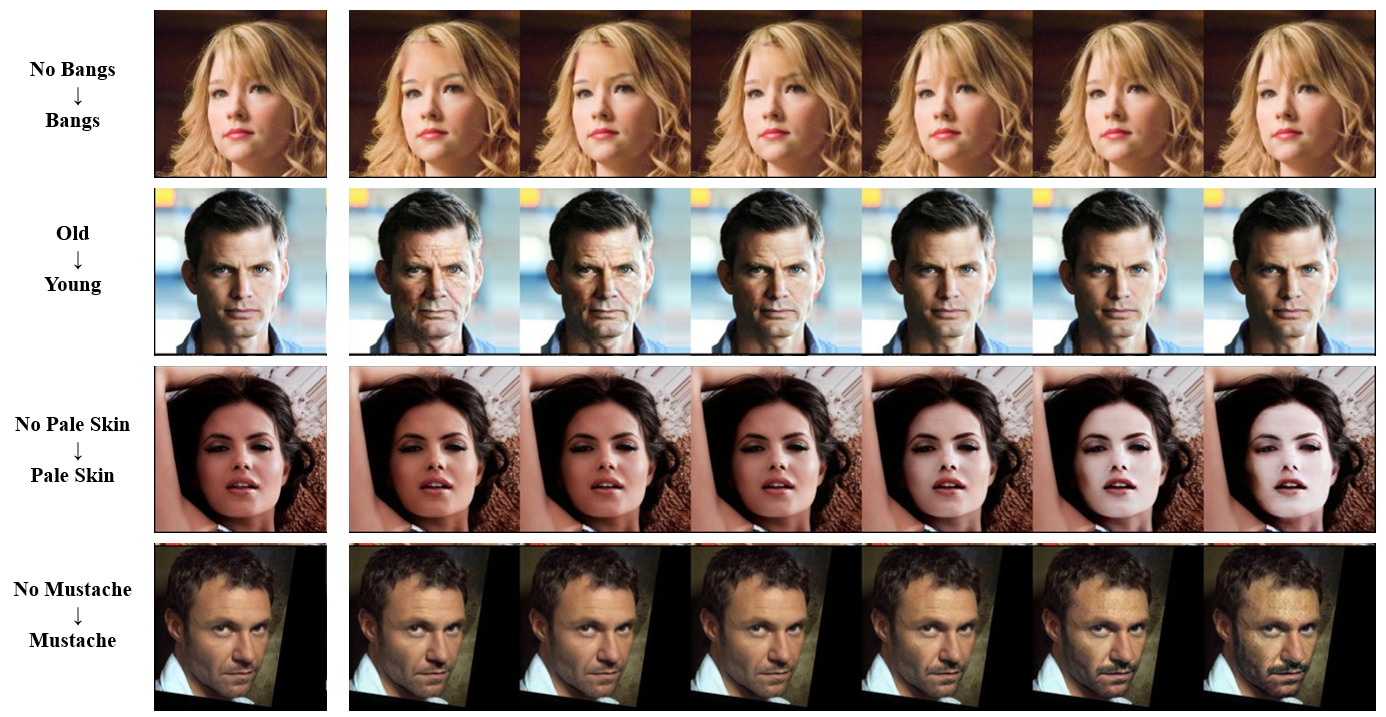}
    \caption{Illustration of attribute intensity control of MagGAN on resolution $256\times256$. The first column is the input image}
    \label{fig:maggan_slide}
\end{figure}

In this section, we show more visual results of MagGAN on resolution $256\times256$. Figure~\ref{fig:maggan_res} shows single-attribute reverse editing, Figure~\ref{fig:maggan_multi} shows multiple-attribute reverse editing, and Figure~\ref{fig:maggan_slide} shows the editing results when attribute intensity varies continuously from 0 to 1.

\begin{table}[t]
\caption{Comparison of different mask-guided reconstruction weight $\lambda_3$ for MagGAN}
\label{tab:l4}
\centerline{
\begin{tabular}{|l||c||c||c||c|c|}
\hline
Methods & \makebox[5em]{MRE $\downarrow$} & \makebox[5em]{FID $\downarrow$} & \makebox[5em]{Avg Acc} & \makebox[5em]{PSNR} & \makebox[5em]{SSIM}  \\ \hline \hline
$\lambda_3 = 0$ & 0.0397 & $1.22$ & $89.6\% $ & $ 39.35 $ & $ 0.980 $ \\ \hline \hline
$\lambda_3 = 100$ & 0.0232 & $ 1.39 $ & $ 85.6\% $ & $ 38.57 $ & $ 0.976 $  \\ \hline
$\lambda_3 = 200$ & 0.0163 & $ \textbf{1.10} $ & $ \textbf{90.0\%} $ & $ \textbf{40.25} $ & $ \textbf{0.984} $ \\ \hline
$\lambda_3 = 400$ & \textbf{0.0157} & $1.33 $ & $ 88.2\% $ & $ 39.34 $ & $ \textbf{0.984} $ \\ \hline
\end{tabular}
}
\vspace{-0.3cm}
\end{table}

\begin{table}[t]
\caption{Comparison of MagGAN with blending trick on resolution $256\times 256$. The blending trick does decrease the mask-aware reconstruction error, but incorporates artifacts at borders, which diminish the visual quality}
\label{tab:blend}
\centerline{
\begin{tabular}{|l||c||c||c||c|c|}
\hline
Methods & {MRE $\downarrow$} & {FID $\downarrow$} & {Avg Acc} & {PSNR} & {SSIM}  \\ \hline \hline
MagGAN & 0.0163& \textbf{1.10}  & \textbf{90.0}\%  & \textbf{40.25}  & \textbf{0.984} \\ \hline
MagGAN w/ blend & \textbf{0.0015} & $1.14$ & $ 83.3\% $ & 37.70 & $ 0.976 $ \\ \hline
\end{tabular}
}
\end{table}

\paragraph{The blending trick to preserve attribute irrelevant regions} In \S~\ref{sec:maskgenerator}, we propose a blending trick to help preserve the attribute-irrelevant regions with alpha composition~\cite{porter1984compositing}. We adopt this blending trick to MagGAN(SP) and report the quantitative results in Table~\ref{tab:blend}. Results show that except MRE reduces significantly, the other metrics are worse than MagGAN when applying the blending trick. From Figure~\ref{fig:maggan_res}, we can also observe that the blending trick generates sharp images, but the visual quality decreases as artifacts are obvious at the boundary of attribute-irrelevant regions.

\begin{figure}
    \centering
    \includegraphics[width=1\linewidth]{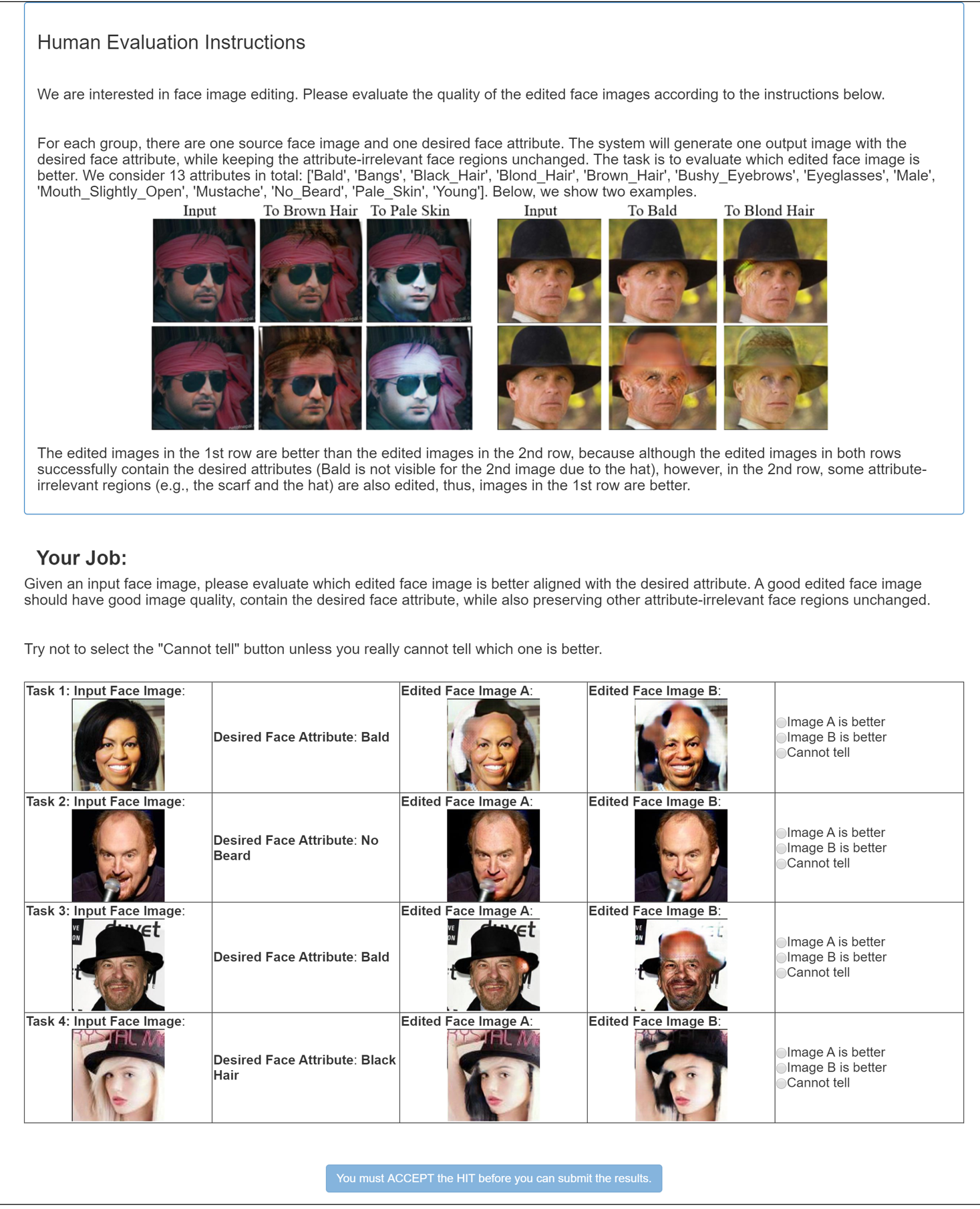}
    \caption{Amazon Mechanical Turk interface of user study. Users are asked to choose the better edited image considering desired attribute}
    \label{fig:user_study}
\end{figure}

\paragraph{User study}  We conduct user study on Amazon Mechanical Turk to compare the generation quality of STGAN and MagGAN. Figure~\ref{fig:user_study} shows the web interface of our user study experiment. 100 input samples are randomly chosen from test set, 50 samples with hat or scarf and 50 samples without. For each sample, 5 attribute editing tasks are performed by STGAN and MagGAN (500 comparison pairs in total). All 5 tasks are randomly chosen from 13 attributes, for subjects with hat, we increase the chance to select hair related attributes, \eg, ``Blonde Hair", ``Bald". The users are instructed to choose the best result which changes the attribute more successfully considering image quality and identity preservation. The user interface also provides a neutral option, which can be selected if the turker thinks both outputs are equally good. To avoid human bias, each sample pair is evaluated by 3 volunteers, thus we have 1500 comparison pairs in total. Only workers with a task approval rate greater than 95\% can participate the study.

\begin{figure}
    \centering
    \includegraphics[width=0.68\linewidth]{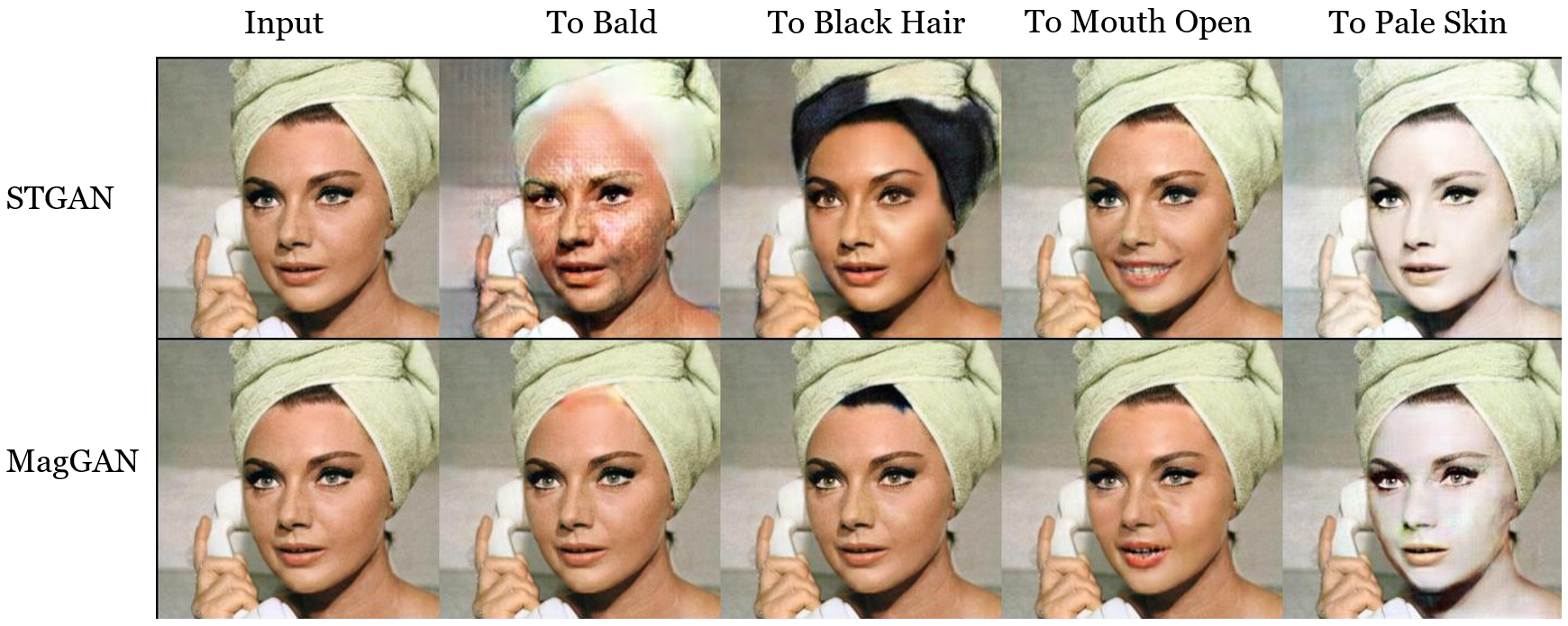}
    \includegraphics[width=0.68\linewidth]{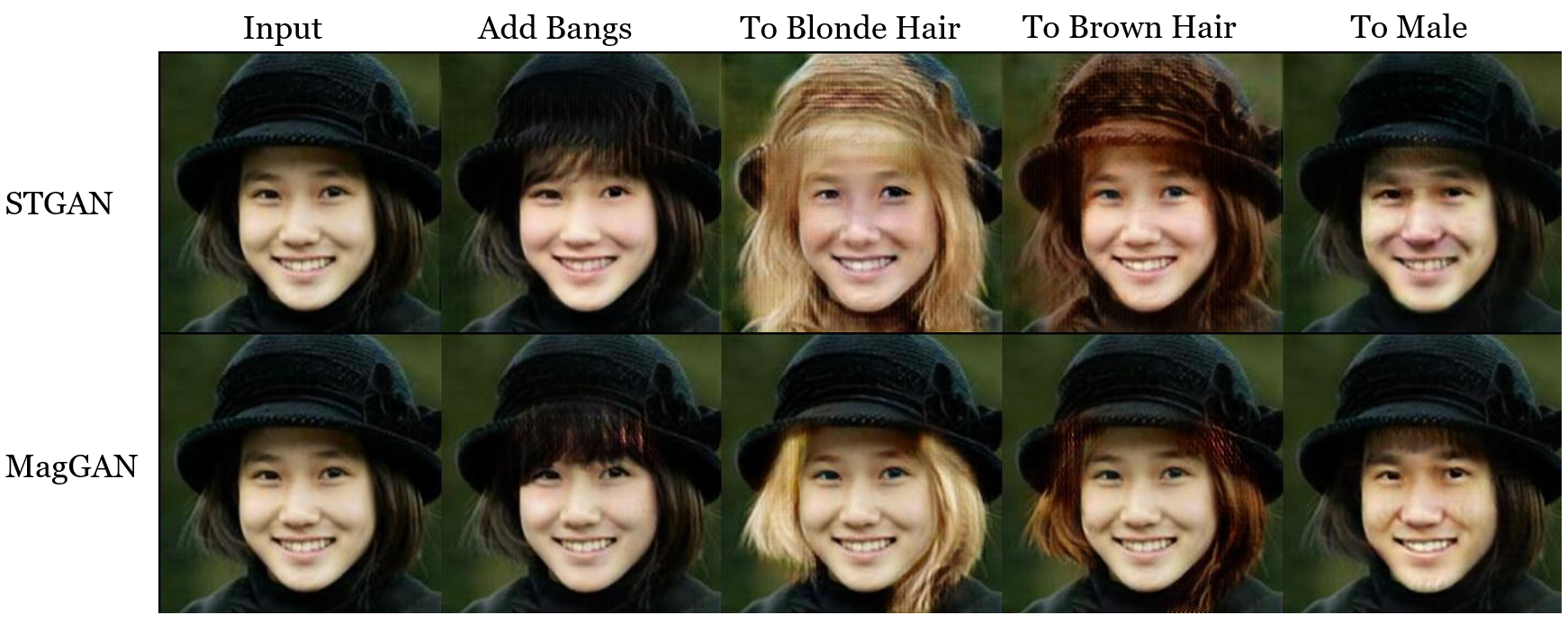}
    \includegraphics[width=0.68\linewidth]{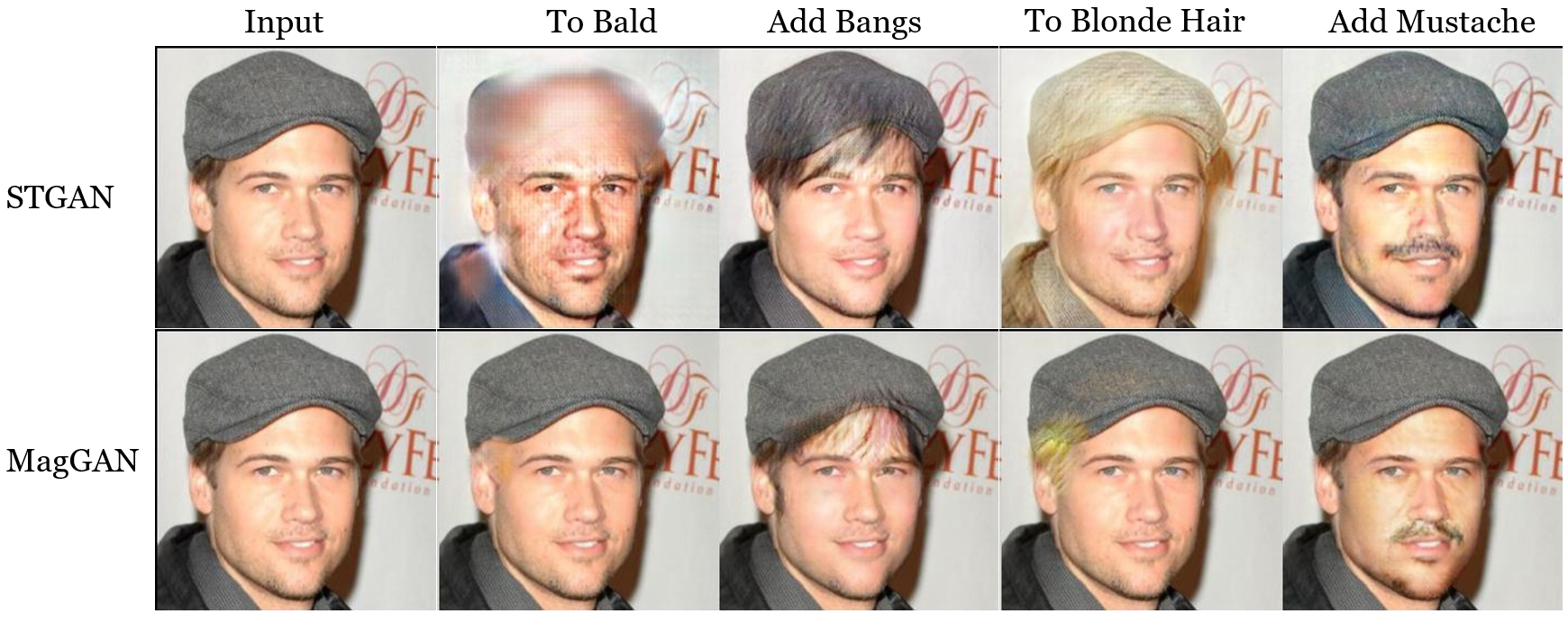}
    \includegraphics[width=0.68\linewidth]{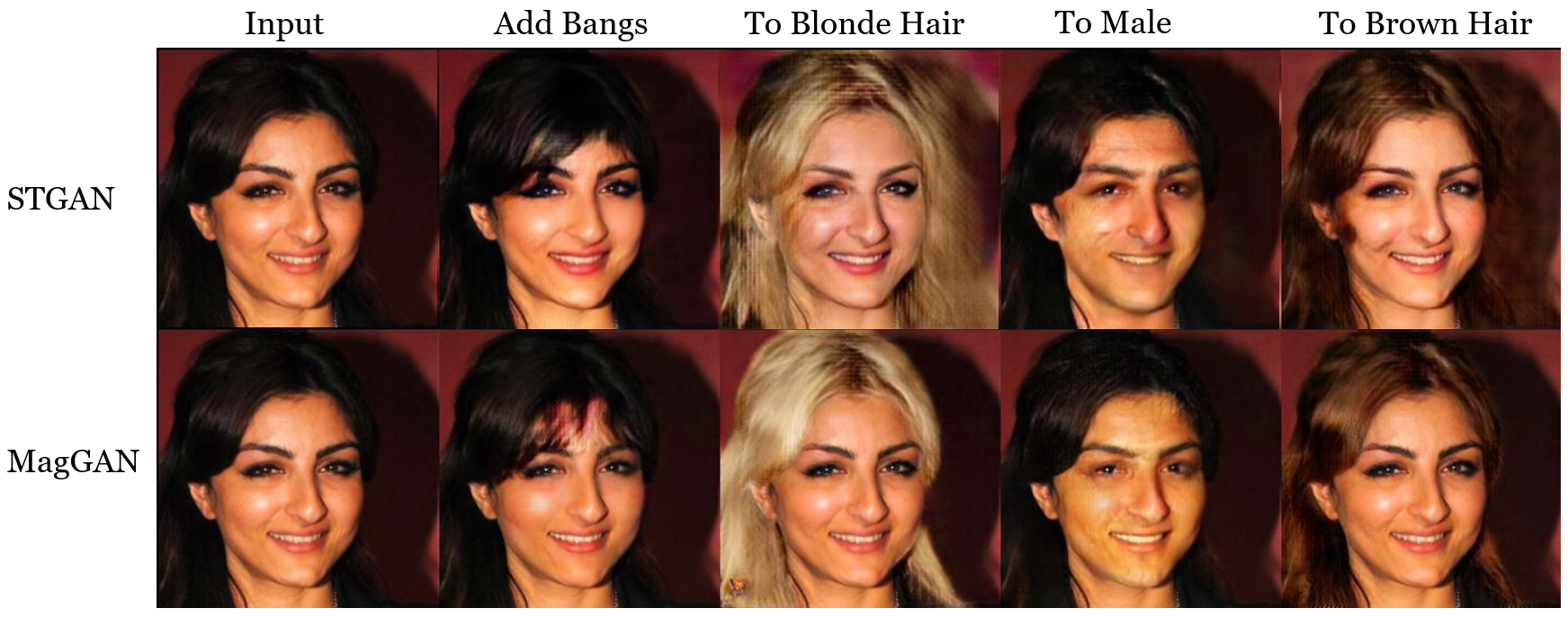}
    \includegraphics[width=0.68\linewidth]{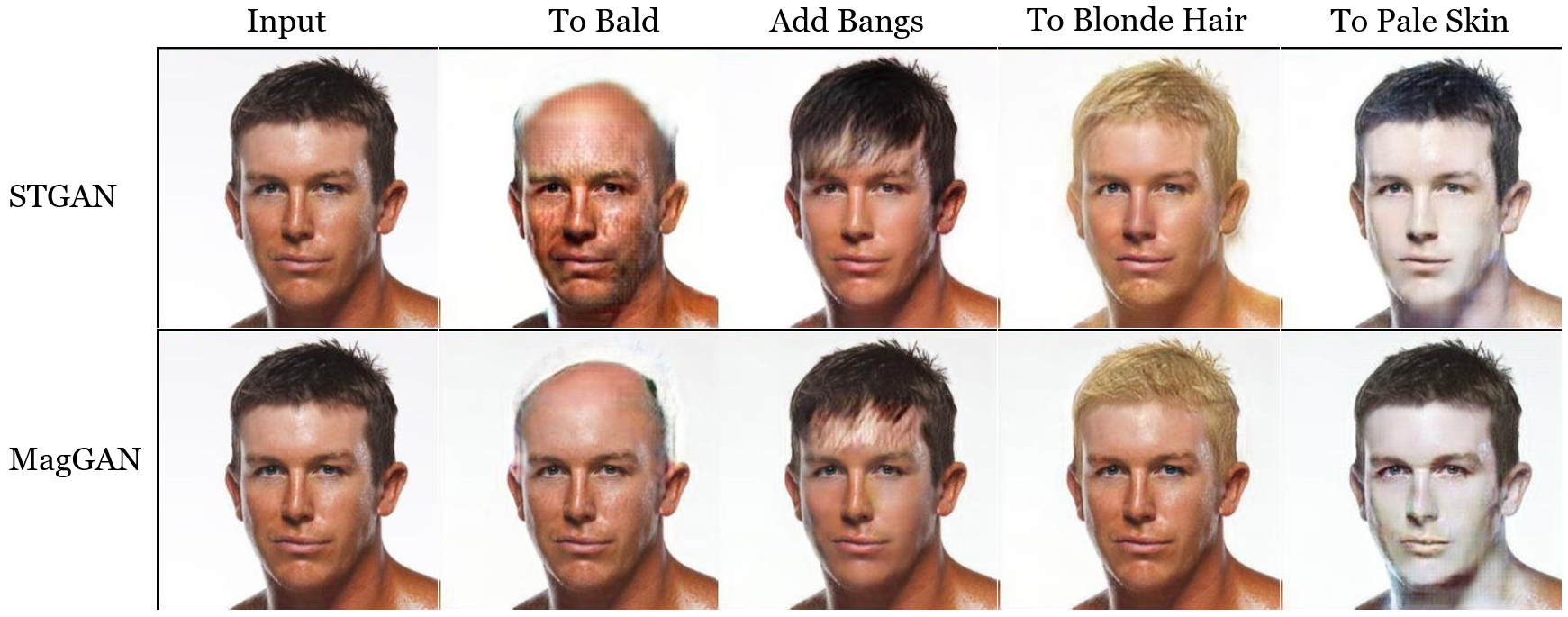}
    \includegraphics[width=0.68\linewidth]{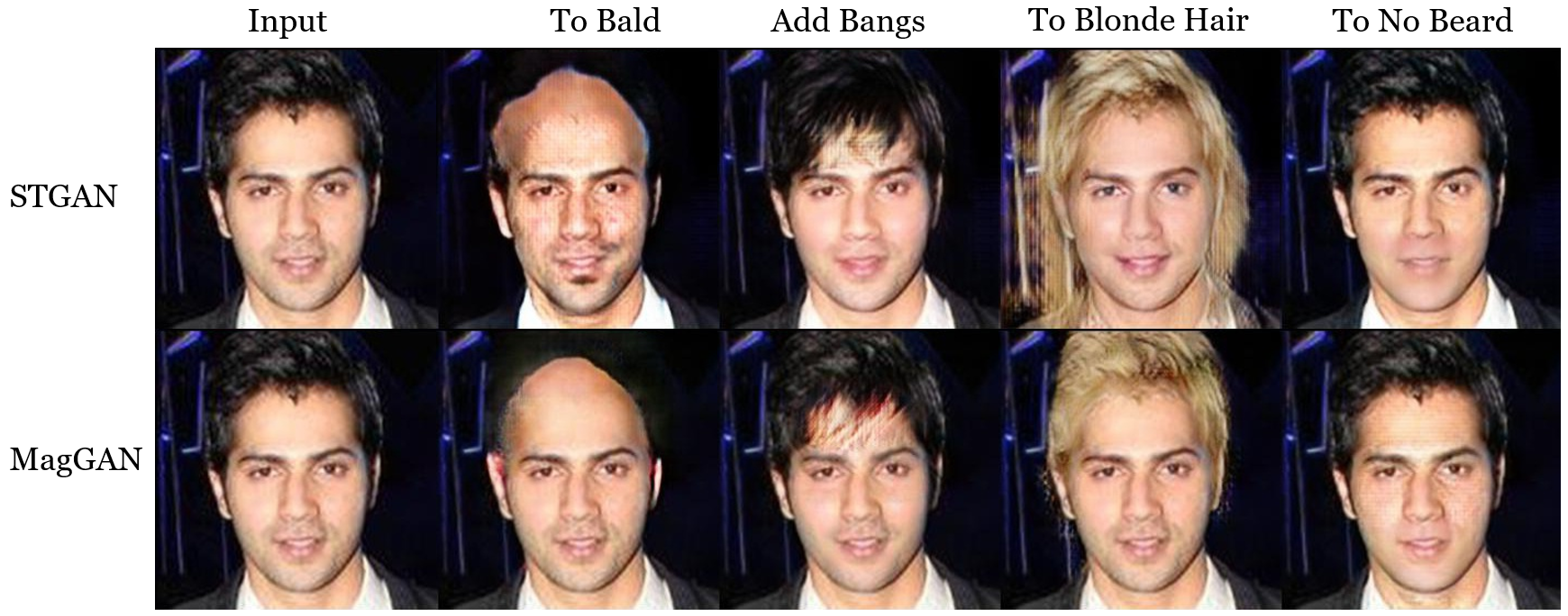}
    \vspace{-3mm}
     \caption{Visual examples of MagGAN and STGAN for user study}
    \label{fig:us_results}
\end{figure}

Figure~\ref{fig:us_results} shows some example visual results for MagGAN and STGAN. Top 3 rows are samples wearing hat or scarf, the last 3 rows are samples without hat. From our observation, MagGAN works better on preserving hat or background regions for with-hat samples, the editing quality also improves for without-hat samples due to the help of mask information. In Table 2 of our submission, we find the gap between MagGAN and STGAN on with-hat samples is not significantly large. We made a meticulous investigation on the collected user study results. We find that the user may misunderstand our instructions by choosing the model with more obvious editing results. For example, the 3rd row in Figure~\ref{fig:us_results}, STGAN achieve more obvious change on ``To Bald" and ``To Blonde Hair" attributes, but in fact, STGAN changes the hat regions which should stay intact. In that situation, MagGAN should be considered as the better model. Thus, such ``failure cases" decrease the votes to MagGAN.

\clearpage
\subsection{Face Attribute Editing on High Resolution}
\label{sec:res_high}

\begin{figure}[t]
    \centering
    \includegraphics[width=0.9\linewidth]{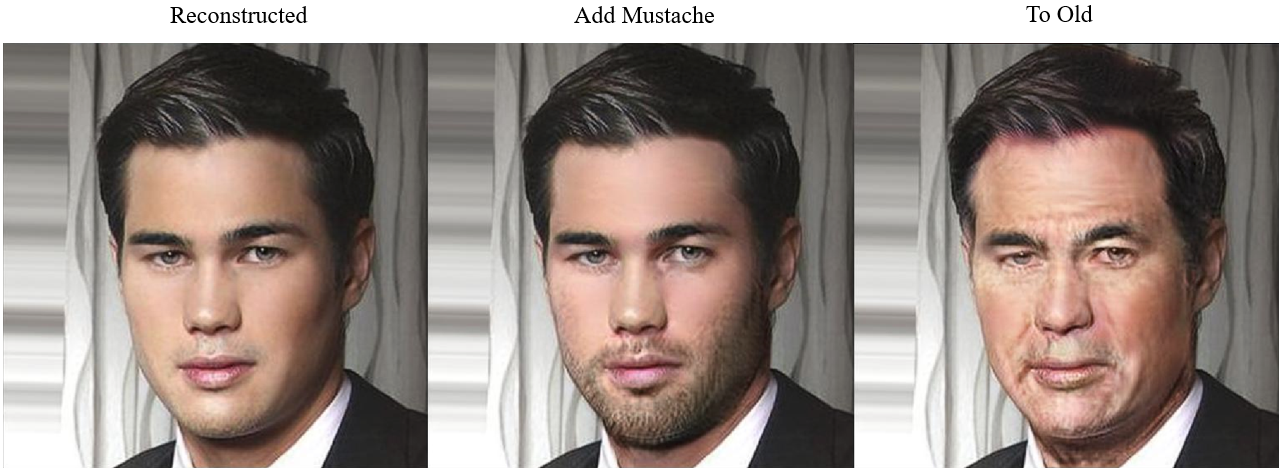}
    \includegraphics[width=0.9\linewidth]{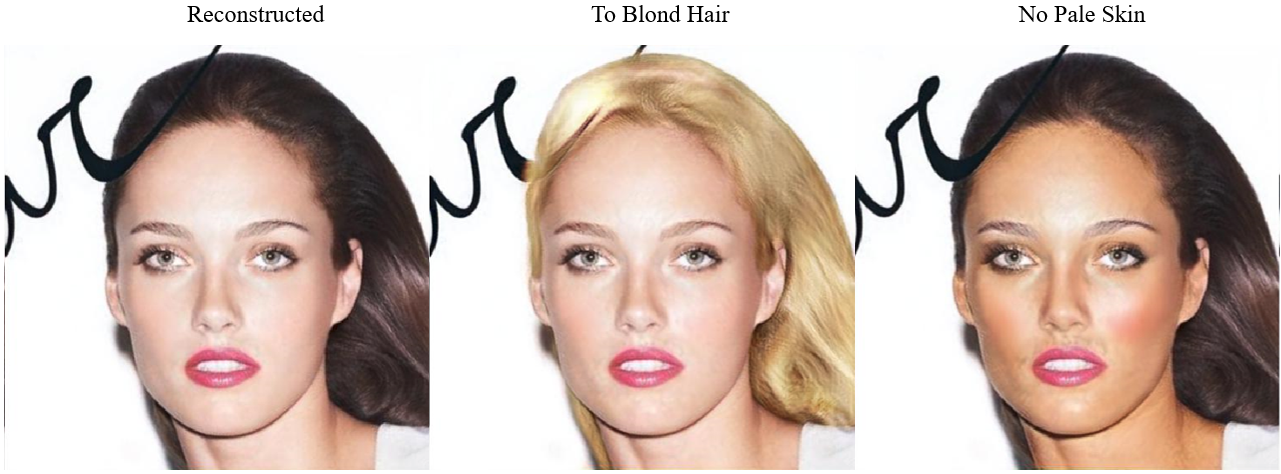}
    \includegraphics[width=0.9\linewidth]{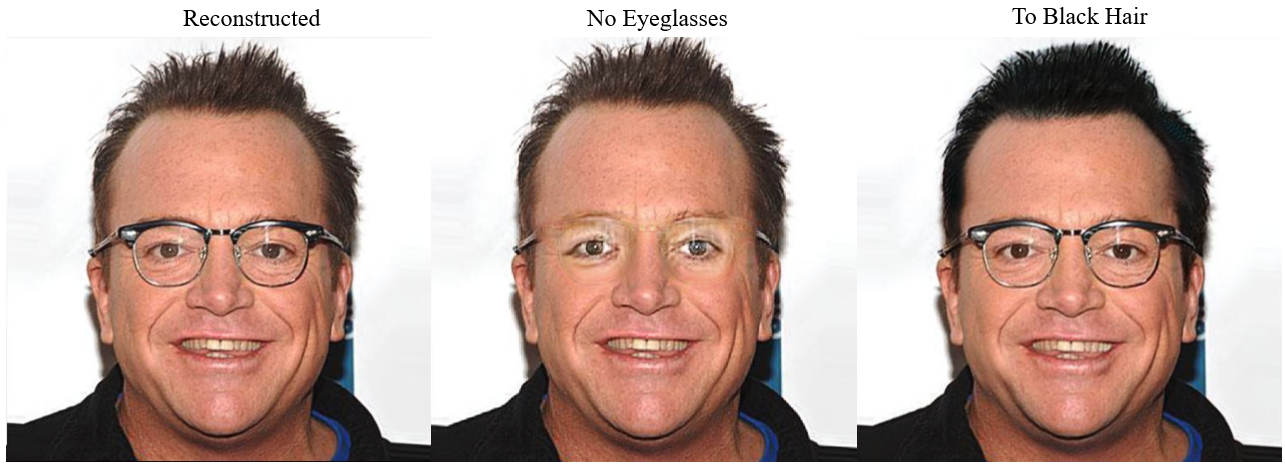}
    \includegraphics[width=0.9\linewidth]{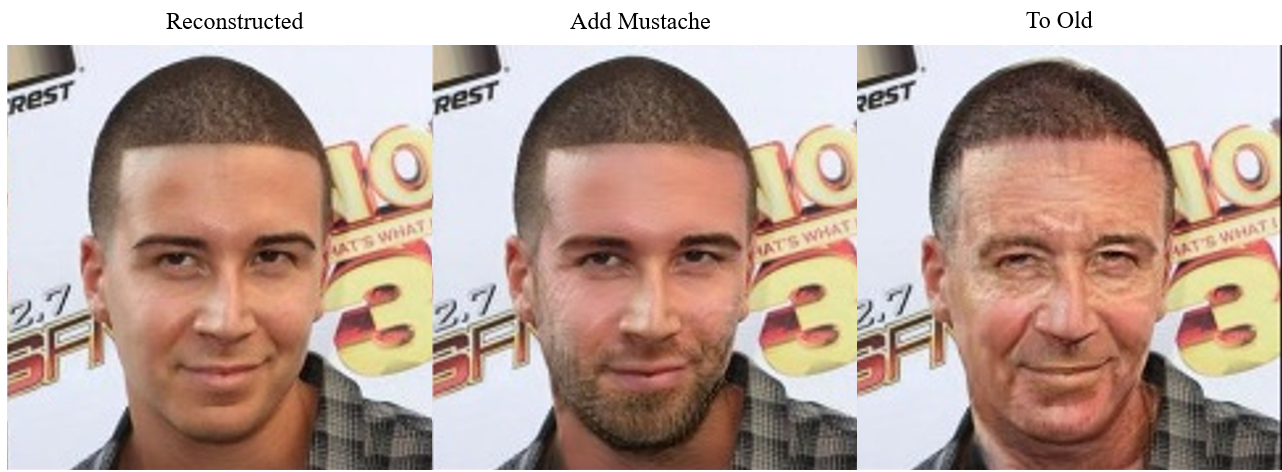}
    \caption{Visual results of MagGAN (using PatchGAN discriminator) on resolution $512\times512$}
    \label{fig:res512}
\end{figure}

We provide more visual results of high-resolution image editing in Figure~\ref{fig:res1024} and Figure~\ref{fig:res512}, for resolution  $1024\times1024$ and $512\times512$ respectively. Fine details of hair and skin can be well reconstructed with the help of PatchGAN discriminator.

\begin{figure}[t]
    \centering
    \includegraphics[width=0.96\linewidth]{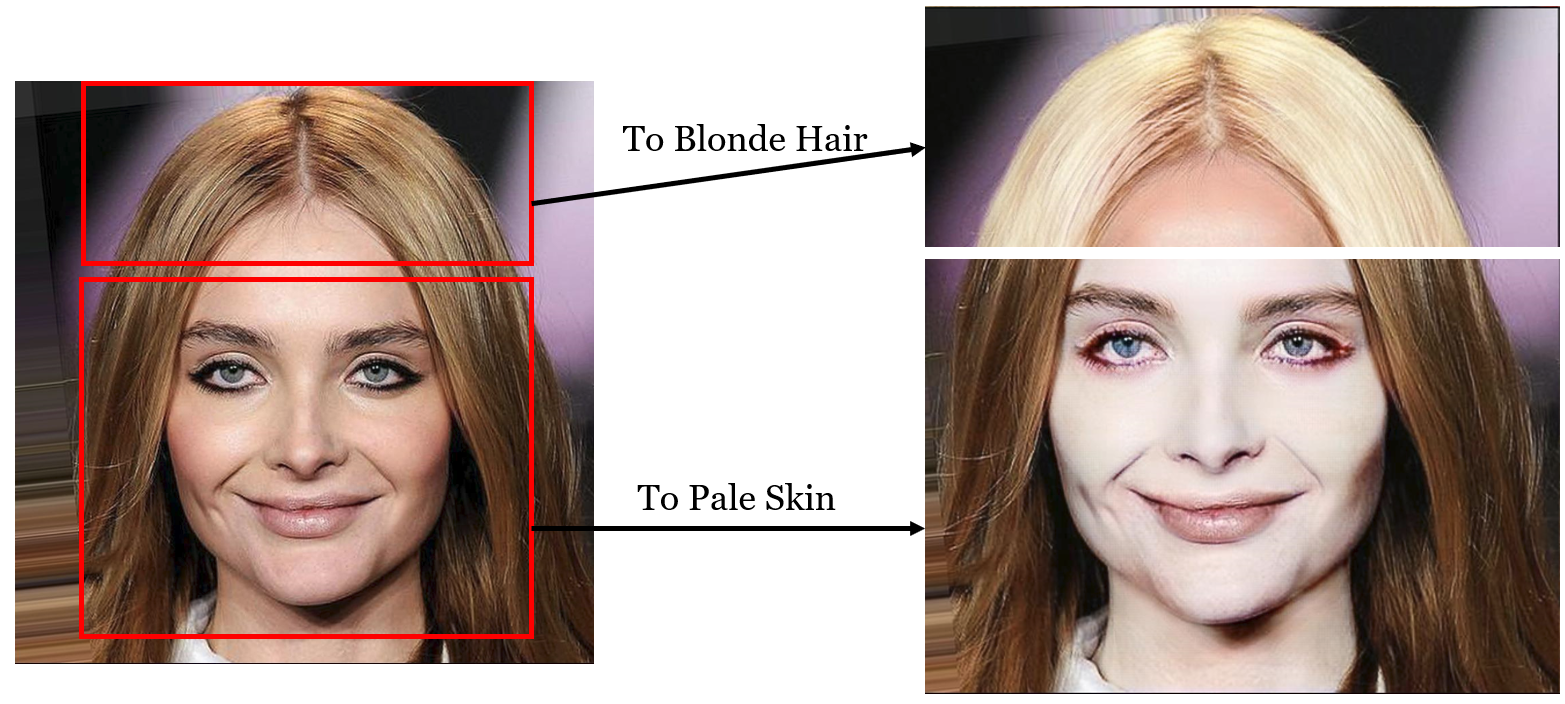}
    \includegraphics[width=0.96\linewidth]{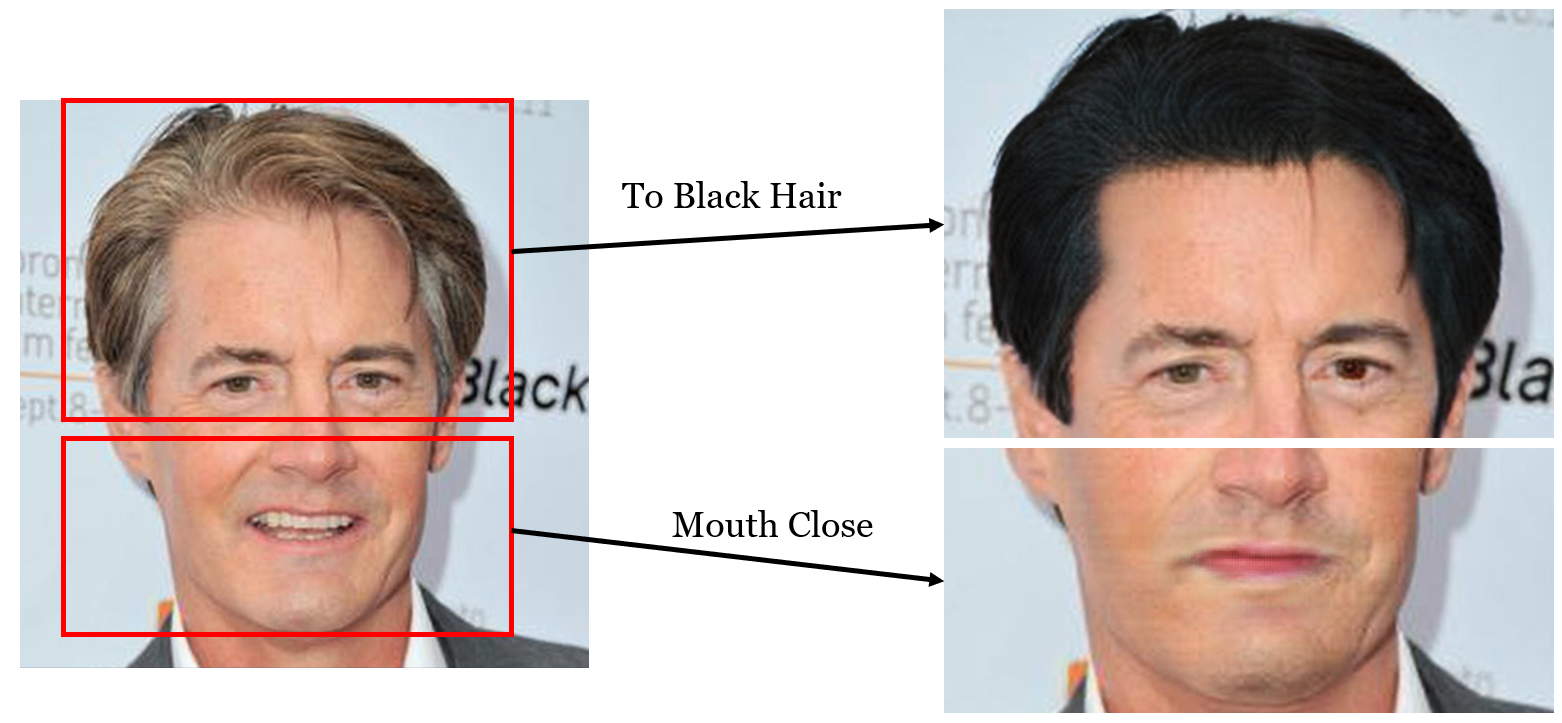}
    \caption{Visual results of MagGAN (using PatchGAN discriminator) on resolution $1024\times1024$. We show the specific sub-regions for better visualization}
    \label{fig:res1024}
\end{figure}

\subsection{Definition of attribute-facial part relationship matrix}
\label{sec:ar_matrix}
We find that facial attributes have strong semantic relationship with specific facial parts. For example, the attribute of "blonde hair" is highly related to the hair regions which should be modified in the edited image if this attribute changes. That leads to a pre-defined attribute-facial part relationship matrix $\mathbf{AR}$ that denotes the relevant regions of each attribute changes. With the help of $\mathbf{AR}$, the preserved mask $M$ to the attribute difference $\mathbf{att}_{\text{diff}}$ can be obtained to computed the mask-guided reconstruction loss (in \S~\ref{sec:training}) and mask-guided condition attribute feature (in \S~\ref{sec:maskgenerator}).

\begin{figure}[h]
    \centering
    \includegraphics[width=0.96\linewidth]{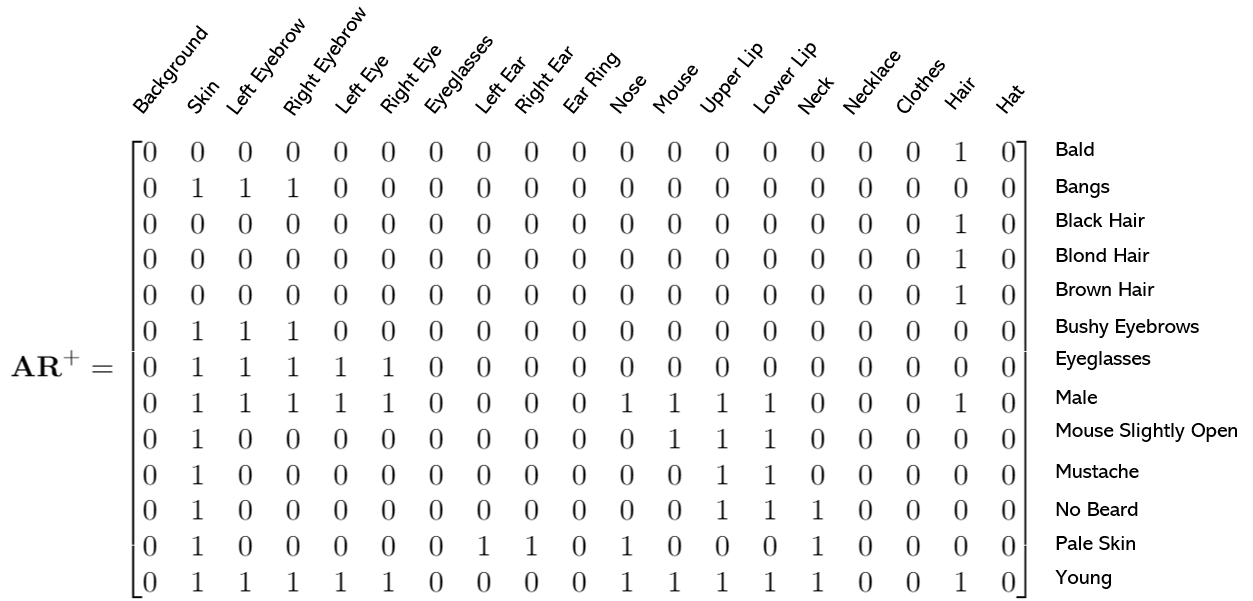}
    \includegraphics[width=0.96\linewidth]{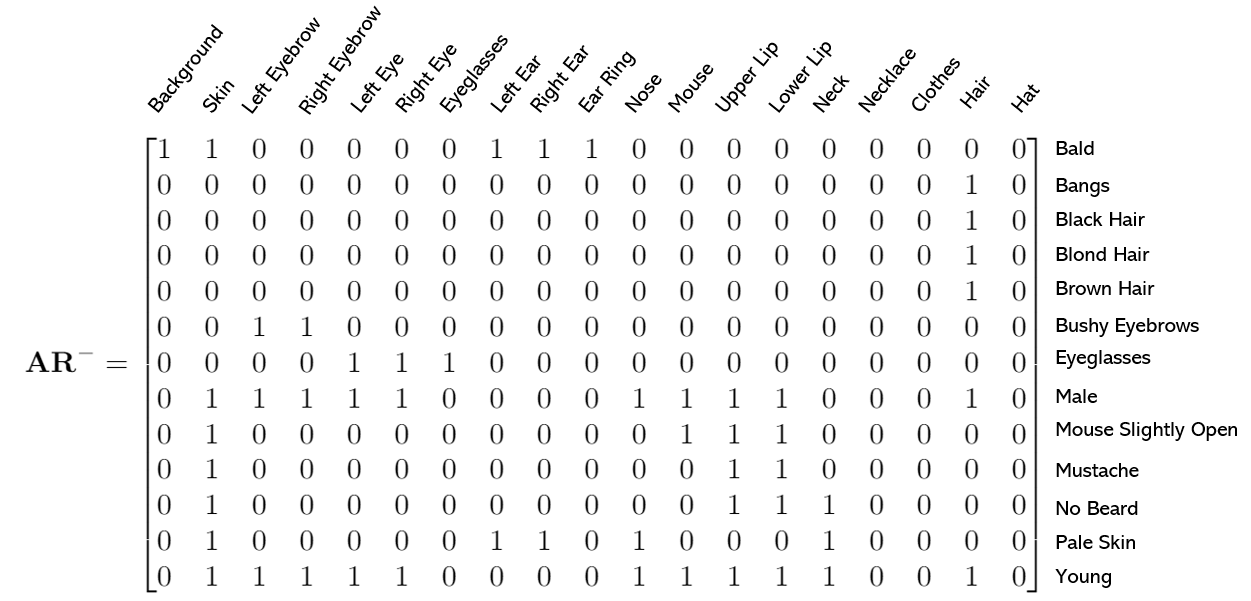}
    \caption{Definition of attribute-part relationship matrices $\mathbf{AR}^{+}, \mathbf{AR}^{-}$. Value 1 represents the attribute and facial part are related, 0 represents that they are irrelevant}
    \label{fig:AR}
\end{figure}

We define two binary attribute-part relation matrices $\mathbf{AR}^{+}, \mathbf{AR}^{-} \in [0,1]^{13\times19}$ in our setting (13 modified attributes and 19 facial parts). We separate the attribute changes to two scenarios: attribute strengthen ($0\rightarrow1$) or attribute weaken ($1\rightarrow0$). The $i$-th row of matrix $\mathbf{AR}^{+}$ or $\mathbf{AR}^{-}$ indicates which facial parts should be modified when the $i$-th attribute is strengthened, \ie, $\mathbf{att}_{\text{diff},i}>0$, or weakened, \ie, $\mathbf{att}_{\text{diff},i}<0$. The detailed definition is in Figure~\ref{fig:AR}.

\subsection{Quantitative Evaluation Metric}
\label{sec:metric}
In \S~\ref{sec:exp}, we apply PSNR (Peak signal-to-noise ratio) and SSIM (Structural Similarity Index) to evaluate the quality of reconstructed images. 

PSNR is most commonly used to measure the quality of reconstruction of lossy compression codecs. In our experiment, we denote the original image as $I$, the reconstructed image as $R$, which takes its original attribute as target attribute. In theory, the input image $I$ and reconstructed image $R$ should be as similar as possible. PSNR (in dB) is defined as:
\begin{equation}
\begin{aligned}
    & \text{PSNR} = 10\cdot\log_{10} (\frac{\text{MAX}^2_I}{\text{MSE}}), \\ 
    & \text{MSE} = \frac{1}{mn}\sum_{i=0}^{m-1}\sum_{j=0}^{n-1} [I(i,j)-R(i,j)]^2
\end{aligned}
\end{equation}
$\mathrm{MAX_I}$ is the maximum pixel value of input image $I$. In general, the larger PSNR value, the better quality the reconstructed image is. 

SSIM (Structural Similarity Index)~\cite{wang2004ssim} is another metric to measure the similarity of image $I$ and image $R$.
The SSIM is defined as:
\begin{equation}
    \text{SSIM}(I, R) = \frac{(2\mu_I\mu_R+c_1)(2\sigma_{IR}+c_2)}{(\mu_I^2+\mu_R^2+c_1)(\sigma_I^2+\sigma_R^2+c_2)}
\end{equation}
where $\mu_I, \mu_R$ denotes the average of $I$ and $R$, $\sigma_I^2, \sigma_R^2$ are the variance of $I$ and $R$, $\sigma_{IR}^2$ denotes the covariance of $I$ and $R$, $c_1, c_2$ are small constants to avoid division instability. Also the larger SSIM value denotes better image quality for reconstructed image.




\clearpage
{\small
\bibliographystyle{splncs04}
\bibliography{references}
}


\end{document}